\documentclass{article} 
\usepackage{iclr2027_conference,times}

\iclrfinalcopy

\usepackage[utf8]{inputenc} 
\usepackage[T1]{fontenc}    
\usepackage{hyperref}       
\hypersetup{hidelinks}
\usepackage{newunicodechar}
\newunicodechar{—}{\textemdash}
\usepackage{url}            
\usepackage{booktabs}       
\usepackage{amsfonts}       
\usepackage{nicefrac}       
\usepackage{microtype}      
\usepackage{xcolor}         
\usepackage{graphicx}       
\usepackage{multirow}
\usepackage{array}
\usepackage{hhline}
\usepackage{amssymb}
\usepackage{booktabs}
\usepackage{tabularx}
\usepackage{makecell}
\usepackage[most]{tcolorbox}
\usepackage{caption}
\usepackage{ragged2e}
\usepackage{makecell}
\usepackage{array}
\usepackage{listings}
\usepackage{amsmath,amssymb}
\usepackage{placeins}
\newcolumntype{Y}{>{\RaggedRight\arraybackslash}X}
\newtcolorbox{promptbox}{
  colback=blue!5,
  colframe=blue!60!black,
  boxrule=0.6pt,
  arc=2pt,
  left=6pt,
  right=6pt,
  top=6pt,
  bottom=6pt,
  fontupper=\ttfamily\small
}
\title{QASM-Eval: A Dataset to Train and\\Evaluate LLMs on OpenQASM-3\\Beyond Quantum Circuits}

\author{Zhenxiao Fu \quad Lei Jiang \quad Fan Chen\\
  Indiana University Bloomington\\
  \texttt{\{zhfu,jiang60,fc7\}@iu.edu}
}

\newlength{\originaltextfloatsep}
\newlength{\originalfloatsep}
\begin{document}

\maketitle
\fancyhead{}

\begin{abstract}
Quantum computing remains in the Noisy Intermediate-Scale Quantum (NISQ) era, with performance constrained by noise. Addressing this limitation requires hardware-facing capabilities beyond gate sequences: mid-circuit measurement and classical feedback for quantum error correction (QEC), precise timing for dynamical decoupling (DD), and pulse-level waveform access for calibration. OpenQASM 3 exposes these capabilities through a hardware-level programming interface. Despite rapid progress in large language models (LLMs) for code generation, datasets targeting these advanced features remain lacking. We introduce QASM-Eval, the first comprehensive dataset to train and evaluate LLMs on OpenQASM 3, targeting completion of hardware-facing constructs within a supplied program context. QASM-Eval comprises an expert-curated test set of 1,200 tasks and a training set of over 4,000 tasks, covering classical logic, timing scheduling, pulse control, and complex tasks inspired by quantum-control workflows. An extended verifier automatically checks syntax, quantum measurement distributions, and program timelines. Our evaluation identifies substantial headroom for OpenQASM 3 code generation and significant gains from targeted fine-tuning. Fine-tuned Llama-3-8B outperforms zero-shot GPT-5.6-Terra, while fine-tuned Llama-3-70B achieves 61.17\% overall pass@1 and approaches few-shot-augmented GPT-5.6-Terra. QASM-Eval provides a benchmark and training resource for specification-guided code generation over the hardware-facing constructs of OpenQASM 3, supporting the development of LLM assistants for quantum programming. Data and code: \url{https://github.com/fuzhenxiao/QASM-Eval}

\end{abstract}

\section{Introduction}
Quantum computing has shown advantage in domains such as chemical simulation~\citep{cao2019QuantumChemReview,lee2023QuantumChemEval}, optimization~\citep{abbas2024QuantumOpt}, and quantum machine learning~\citep{cerezo2022QuantumML,biamonte2017quantumML2}. Yet practical quantum hardware remains in the Noisy Intermediate-Scale Quantum (NISQ) regime~\citep{chen2023RecentNISQ,preskill2018WellCitedNISQ},where quantum processors still suffer from quantum noise, a stochastic disturbance corrupting quantum states and causing computational deviations. Although numerous approaches exist to mitigate the effects of noise, such as quantum error correction (QEC)~\citep{gottesman2010QEC}, dynamical decoupling (DD)~\citep{khodjasteh2005DD}, and routine calibration~\citep{knill2008randomized}, each of them requires specific, low-level hardware controls. First, QEC relies on mid-circuit measurement and runtime classical computation/feedback~\citep{ryan2021RealQEC}. Second, intervention methods such as DD are highly dependent on precise control over gate timing~\citep{bylander2011DD_timing}. Third, because qubits naturally drift over time and calibrated gate fidelities keep degrading~\citep{burnett2019drift1,proctor2020drift2}, sustaining performance requires not only routine calibration but also pulse-level access to actively tune control waveforms. However, existing high-level quantum programming tools such as Qiskit~\citep{javadiabhari2024qiskit}, Cirq~\citep{Cirq2025}, and PennyLane~\citep{bergholm2022pennylane}, lack comprehensive support for these fine-grained hardware controls.

OpenQASM~3 \citep{cross2022openqasm3} addresses these limitations by serving as a hardware-aware intermediate representation that links algorithms to physics. Unlike high-level tools, OpenQASM~3 exposes hardware instructions that directly fulfill the diverse operations required for noise mitigation. First, it supports hardware-embedded classical logic and control flow, enabling runtime mid-circuit operations. Second, it introduces explicit gate timing and scheduling constructs that allow dynamic operation duration, alignment, and delay insertion. Finally, OpenQASM~3 further extends to pulse-level control, allowing developers to describe or tune physical control waveforms directly, thus providing users with a means to actively manage calibration details. Taken together, these capabilities make OpenQASM~3 a key enabler for improving quantum-computing performance in the NISQ era.

Given the increasing complexity and features of OpenQASM~3, as well as the rapid progress of Large Language Models (LLMs) for code generation~\citep{jimenez2023swe,schluntz_2025_ClaudeSWE,nam2024code_understanding,joel2024LLM_coding_domain_specific}, using LLMs to assist OpenQASM programming is a natural next step. However, this direction is currently limited by the lack of appropriate datasets. Existing OpenQASM-related resources fall into two unsatisfactory categories. Some datasets, such as Veri-Q~\citep{chen2022veriq} and QASMBench~\citep{qasmbench}, were designed to benchmark quantum algorithms or hardware platforms rather than to support LLM training or evaluation. Others, such as QCircuitBench~\citep{yang2024qcircuitbench} and Agent-Q~\citep{jern2025agentQ}, remain confined to gate-sequence circuit generation and do not cover the critical features of OpenQASM~3, including classical logic, timing scheduling, and pulse control. As summarized in Table~\ref{tab:resource-comparison}, no existing resource simultaneously targets LLM-based code generation and captures the core features that define OpenQASM~3's central role in improving NISQ-era quantum-computing performance. 

\begin{table}[!htbp]
  \centering
  \caption{Comparison of representative OpenQASM-related resources by whether they target LLM code generation and whether they cover major OpenQASM~3 feature dimensions.}
  \label{tab:resource-comparison}
  \small
  \setlength{\tabcolsep}{3.5pt}
  \begin{tabular}{lcccc}
    \toprule
    \textbf{OpenQASM datasets} & \textbf{LLM-targeted} & \textbf{Classical Logic} & \textbf{Timing Scheduling} & \textbf{Pulse Control} \\
    \midrule
    \textbf{Our work (QASM-Eval)} & $\checkmark$ & $\checkmark$ & $\checkmark$ & $\checkmark$ \\
    QCircuitBench~\citep{yang2024qcircuitbench} & $\checkmark$ & $\times$ & $\times$ & $\times$ \\
    Agent-Q~\citep{jern2025agentQ} & $\checkmark$ & $\times$ & $\times$ & $\times$ \\
    Veri-Q~\citep{chen2022veriq} & $\times$ & $\times$ & $\times$ & $\times$ \\
    QASMBench~\citep{qasmbench} & $\times$ &$\times$ & $\times$ & $\times$ \\
    \bottomrule
  \end{tabular}
\end{table}

To address these gaps, we propose \textbf{QASM-Eval}: paired specification--completion tasks for OpenQASM~3, with fixed program backgrounds and an executable evaluation interface. Hardware-facing constructs define the coverage; observable outputs and schedules define the evaluation. Our main contributions are:
\begin{itemize}
  \item We construct an OpenQASM~3 task suite for both LLM training and evaluation. We release a test set with 1200 tasks, a training set with 4000+ tasks, and two targeted fine-tuned models. The dataset covers key OpenQASM~3 capabilities, including classical logic/control flow, timing constraints, and pulse-level control. To validate these features, we extend existing toolchains with new OpenQASM~3 support so that generated code can be automatically verified for syntactic, semantic, and scheduling constraints.
  \item Using the QASM-Eval test set, we characterize current OpenQASM~3 capabilities and evaluate the effectiveness of our training data: QASM-Eval fine-tuning improves pass@1 by 36.95--45.47 percentage points for Llama-8B and Llama-70B; the fine-tuned 8B model exceeds zero-shot GPT-5.6-Terra, while the fine-tuned 70B model approaches few-shot-augmented GPT-5.6-Terra.
\end{itemize}

\section{Related Work}
\paragraph{Quantum Computing in the NISQ Era}
Practical quantum computing remains constrained by the NISQ regime, where device performance is limited by quantum noise originating from various sources, such as state-preparation and measurement (SPAM) errors~\citep{ryan2021RealQEC}, imperfect gate implementations~\citep{rudinger2019gateError}, crosstalk~\citep{ahsan2022crosstalkerror} and decoherence~\citep{krantz2019coherencetime}, which accumulate as circuits deepen. A broad line of work therefore focuses on noise mitigation and suppression. Quantum error correction (QEC) \citep{gottesman2010QEC,ryan2021RealQEC} provides a principled route to fault tolerance by encoding logical qubits into multiple physical qubits, checking the status of qubit via mid-circuit measurement, and locating potential errors via classical computation~\citep{ashikhmin2014syndrome,bhatnagar2023syndrome2}. Dynamical decoupling (DD) \citep{khodjasteh2005DD,bylander2011DD_timing} reduces noise by inserting carefully timed pulse sequences to average out low-frequency noise, which makes it highly sensitive to precise control over operation timing and spacing, especially in dynamic circuits. Calibration is also critical. Although providers routinely recalibrate devices to maintain fidelity~\citep{knill2008randomized}, qubit frequencies, gate parameters, and readout characteristics still keep drifting between calibrations~\citep{burnett2019drift1,proctor2020drift2}. Therefore, advanced users may additionally require direct access to customized control waveforms~\citep{khaneja2005DMRpulse,motzoi2009DRAGpulse}. Yet these hardware-facing requirements are only weakly exposed in most high-level quantum software stacks.

\paragraph{OpenQASM~3 Language}
OpenQASM~3 \citep{cross2022openqasm3} was introduced to bridge this gap between high-level program specification and low-level hardware execution. Its most important advance over earlier circuit-description formats is support for embedded classical computation and control flow, including branching and runtime decisions based on measurement outcomes, which makes mid-circuit adaptive protocols expressible within the language itself. OpenQASM~3 also incorporates explicit timing and scheduling constructs, allowing programmers to control operation duration, alignment, and inserted delays with much greater precision; this is essential for expressing temporally sensitive techniques such as DD or hardware-aware gate orchestration. In addition, the language extends toward pulse-level control, enabling the specification and tuning of physical control waveforms needed for calibration-sensitive experiments and custom hardware manipulation. These features position OpenQASM~3 as a practical interface for NISQ-era programs that must interact closely with device physics rather than remain at the level of abstract circuits.

\paragraph{OpenQASM Datasets}
The application of LLMs to quantum programming is an emerging field, and current datasets reflect a strong bias toward high-level SDK ecosystems rather than intermediate representations. Datasets such as QDataset \citep{perrier2022qdataset}, Qiskit-HumanEval \citep{vishwakarma2024qiskit}, QuanBench \citep{guo2025quanbench}, and MQTBench \citep{quetschlich2023mqt} primarily evaluate host-side Python code generation. Among resources targeting OpenQASM directly, Veri-Q \citep{chen2022veriq} and QASMBench \citep{qasmbench} focus on compiler optimization and hardware benchmarking using static circuit files. QCircuitBench \citep{yang2024qcircuitbench} and Agent-Q~\citep{jern2025agentQ} are the closest antecedents for LLM research, as QCircuitBench pairs circuits with natural-language descriptions and Agent-Q includes various circuits designed specifically for optimization problems; however, they remain restricted to basic, gate-sequence scripts. Consequently, the literature currently lacks any benchmark that captures the dynamic, hardware-facing features of OpenQASM~3—specifically classical logic, explicit scheduling, and pulse control.

\section{QASM-Eval Dataset}
In this section, we introduce QASM-Eval. To our knowledge,  QASM-Eval is the first dataset designed for LLMs that targets OpenQASM~3 and its advanced hardware-level features beyond specific quantum circuits. The test set contains 1200 OpenQASM~3 quantum-coding tasks spanning diverse themes, while the training set can be generated at scale (we generate 4,000 tasks for fine-tuning in this work, and the generation pipeline supports further scaling), together with a corresponding simulation-based testbed. We elaborate two key aspects: (1) To comprehensively cover the new features introduced by OpenQASM~3, our dataset has three major task categories including classical logic, timing scheduling, and pulse control, as well as one extra category of challenging complex tasks that integrate all three categories based on realistic applications, such as QEC, DD and calibration (2) To support large-scale data generation while preserving correctness, we adopt a dataset-construction pipeline that combines curated templates, LLM-assisted augmentation, and expert review, a strategy that has proven effective in prior LLM-dataset works~\citep{xie2025pipelineexample1,lu2025pipelineexample2,joseph2025pipelineexample3}.

\subsection{Task Category}
QASM-Eval comprises four task categories as listed in Table \ref{tab:task-categories}. Three categories target core new capabilities in OpenQASM~3: (1) \emph{classical-logic} tasks that exercise classical control and computation, (2) \emph{timing scheduling} tasks that focus on timing and scheduling primitives, and (3) \emph{pulse control} tasks that involve low-level pulses, calibrations, and related functionality. In addition, we include (4) \emph{complex} category that composes all features into more challenging, real-world problems such as QEC, DD and calibration. Further details can be found in Appendix \ref{appendix:datasetdetails}.

\begin{table}[!htbp]
\centering

\caption{Task taxonomy of QASM-Eval with four categories. Each category contains 300 tasks for testing and 1000+ tasks for training}

\small
\setlength{\tabcolsep}{4pt}
\renewcommand{\arraystretch}{1.0}
\begin{tabularx}{\columnwidth}{|l| c|c| Y|}
\toprule
\textbf{category} & \textbf{\# test}& \textbf{\# train} & \textbf{involved features/functionalities} \\

\midrule
classical & 300 & 1000+ &
if/else, mid-circuit measurement, while loop, for loop, switch statement, arithmetic calculation, dynamic unit, dynamic data type, type casting, array, dynamic comparison, bit-wise operation, external functions \\
\hline
timing & 300 & 1000+ &
delay, duration, hybrid units, stretch, multiple stretch, alignment, proportional arrangement, dynamic duration, box operations, barrier \\
\hline
pulse & 300 & 1000+ &
wave calibrate/rewrite/measure, shift phase, modulation, custom waveforms, frame sync, param gate, multiplex readout, phase tracking \\
\hline

complex & 300 & 1000+ &
all of above + real-world application scenarios including QEC, DD, calibration, RAMSEY, Hahn echo, parity check, crosstalk detection, \dots \\
\bottomrule

\end{tabularx}
\label{tab:task-categories}
\end{table}

\paragraph{Classical Logic Tasks}
This category places quantum circuits within an executable classical-control framework. Tasks typically require using conditional branches to drive quantum operations, or computing over and iterating through classical objects to decide subsequent circuit behavior. For instance, a task may compare mid-circuit measurement outcomes against classical data to enter an \texttt{if}/\texttt{else} branch, or use \texttt{while} loops to repeatedly probe or steer qubit states, with \texttt{break} and \texttt{continue} to constrain real-time control flow. We also incorporate standard classical instructions such as Boolean comparisons, bitwise operations (including shifts and \texttt{popcount}), and type casting to ensure that each problem exercises a nontrivial coupling of classical logic and quantum execution.

\paragraph{Timing Scheduling Tasks}
This category focuses on program-level scheduling of quantum operations. Basic tasks require explicitly inserting \texttt{delay} instructions and using the \texttt{duration} type to compose timing constants and arithmetic at compile time. Tasks may also use \texttt{dt}, a backend-defined unit tied to the sampling period, to avoid mismatches between waveform sampling rates and time resolution. More advanced tasks involve \emph{stretchable} time intervals whose values are solved by the compiler without relying on specific calibration lengths, enabling constraints such as alignment, uniform spacing, or ``as-late-as-possible'' placement. For example, combining \texttt{stretch} with \texttt{barrier} can enforce that selected operations complete simultaneously.

\paragraph{Pulse Control Tasks}
This category targets low-level, executable pulse programs. Tasks require manipulating waveforms under explicit resource constraints, selecting the correct control targets, initializing sequence objects, executing pulses, and performing readout. For example, they exercise \emph{ports} as hardware-facing I/O abstractions and \emph{frames} as stateful containers that track clocking, carrier phase, and frequency. Programs update these states via instructions such as \texttt{set\_phase}, \texttt{shift\_phase}, and \texttt{set\_frequency}, and then schedule a waveform onto a frame using \texttt{play}. We further include composition operators such as \texttt{mix}, \texttt{sum}, \texttt{phase\_shift}, and \texttt{scale} to cover modulation, phase tracking, custom waveform synthesis, and multiplexed readout.

\paragraph{Complex Tasks} 
This category composes at least two of the three categories above (classical logic, timing, and pulse control) to approximate realistic experimental workflows, including typical scenarios such as QEC, DD, and calibration. Representative tasks include \texttt{syndrome\_feedforward\_idle\_scheduling} for QEC, which integrates pulse-level syndrome capture with conditional feedforward logic and precise idle padding (covers classical logic and pulse control). Other examples are \texttt{boxed\_dynamic\_decoupling} for DD, which embeds a decoupling pulse sequence within a fixed time window using elastic gaps (covers timing and pulse control), and \texttt{measurement\_crosstalk\_calibration} for calibration, which evaluates hardware crosstalk by interleaving simultaneous driving, readout, and conditional bitwise operations (covers all three categories). These tasks involve a broader set of interacting features and are therefore well-suited for evaluating LLM performance in practice-oriented quantum programming settings.

\subsection{Dataset Construction}
\begin{figure}[!t]
  \centering
  \includegraphics[width=\linewidth,height=0.35\textheight,keepaspectratio]{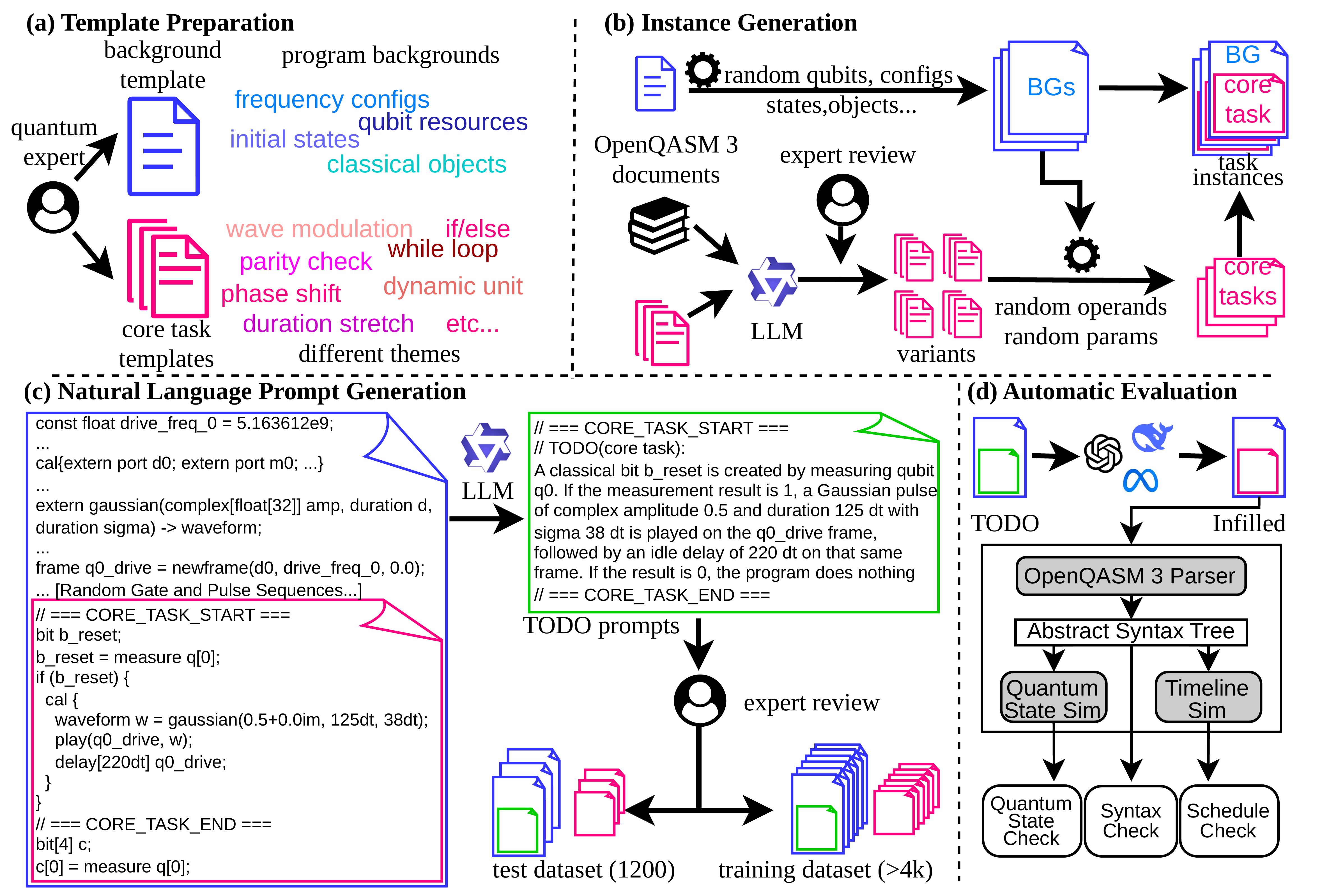}
  \caption{Overview of the QASM-Eval construction pipeline.}
  \label{fig:construction-pipeline}
\end{figure}
\paragraph{Template Preparation} Given the templated nature of quantum programs\citep{qasmbench}, we follow the template-based approach used in QCircuitBench~\citep{yang2024qcircuitbench}. To ensure both the diversity and correctness of QASM-Eval, we first ask quantum-programming experts to design two types of templates(Figure~\ref{fig:construction-pipeline}-a). The background templates randomize the circuit context (e.g., qubit configurations, default frequencies, etc.) and applies random gate sequences to diversify the initial quantum state. The core-task templates target specific themes like \texttt{if}/\texttt{else} control flows or pulse operations, and leaves placeholders for operands and parameters for randomized selection later. 
\paragraph{Instance Generation}
Instance generation process is shown in Figure~\ref{fig:construction-pipeline}-b. Based on templates, we generate randomized background circuits (BGs) with diverse context such as qubit configurations. Then, to further increase diversity of core tasks, we use coder-LLM (Qwen3-Coder-480B) augmented with OpenQASM~3 documentation to create stylistic and structural variants for each core task template (see examples in Appendix~\ref{appendix:datasetexamples}). All template variants were reviewed by experts to ensure correctness. Conditioned on BGs, the core generator selects appropriate objects, operands and parameters, instantiating a theme-specific core-task template (including variants) to produce a complete OpenQASM~3 program. We also append an explicit observation step (qubit measurements) to ensure task outcomes are well-defined and verifiable. 
\paragraph{Natural Language Prompt Generation}
Finally, we assemble the dataset into specification--solution pairs (Figure~\ref{fig:construction-pipeline}-c). Specifically, we merge a sampled background with one core task to form a complete OpenQASM program. We then use LLM to generate a natural-language prompt describing the core task, embedding it into the program as a TODO comment. This yields the final paired data: (i) the specification (background + TODO comment) and (ii) the reference solution (background + core-task code). 
\paragraph{Validation}
Every generated instance is checked with the same syntax, state-distribution, and scheduling machinery used during evaluation; instances whose reference programs do not satisfy the intended behavior are excluded. To prevent leakage in the theme-held-out experiments, each fold's training data exclude all themes assigned to that fold's test partition. In addition, experts independently reviewed 200 test problems, and spot-checked 400 training instances, checking for ambiguity, specification errors, and answer leakage.

\subsection{Evaluation Method} \label{subsec: evalmethod}
We evaluate model-generated solutions with an automated verifier that checks both \emph{syntax} and \emph{semantics}. Semantic verification is further decomposed into two complementary views: the \emph{observable output distribution} induced by the program and the \emph{execution schedule} implied by timing primitives. To support these checks under OpenQASM~3 features that are not fully covered by existing toolchains~\citep{ibm_openqasm3_feature_table}, we implement a quantum-simulation-based automatic verifier as shown in Figure~\ref{fig:construction-pipeline}-d (details in Appendix~\ref{appendix:verifierarchitecture}).

\paragraph{Syntax checking.}
We parse each submitted program using the official OpenQASM parser and obtain an abstract syntax tree (AST). The verifier inspects statement types and their constituents to ensure that the program is well-formed and adheres to the OpenQASM~3 syntax. In addition, we explicitly check whether the construct elements (e.g., a \texttt{while} loop or \texttt{switch} statement) required by the task \texttt{TODO} prompt are present in the submitted solution. Failed samples are categorized as \textit{syntax errors} or \textit{element errors}.

\paragraph{Quantum-state checking.}
Because common frameworks such as Qiskit~\citep{javadiabhari2024qiskit} and Cirq~\citep{Cirq2025} only support a subset of OpenQASM~3 functionality, we first extend our frontend with new OpenQASM~3 semantics, and then use statevector simulation (built on Qiskit) and pulse-level simulation (built on QuTiP~\citep{lambert2026qutip}) to compute the final quantum state. We compare the joint distribution of simulated measurement outcomes against the reference behavior defined by the task. Failed samples are categorized as \textit{distribution errors}.

\paragraph{Schedule checking.}
To validate timing behavior, we assign default durations to supported OpenQASM~3 operations and expose an interface to override them when needed. After interpreting OpenQASM~3 timing constructs, the verifier computes a software schedule and produces a virtual timeline, which is then checked against the task's scheduling constraints. Failed samples are categorized as \textit{timeline errors}.

\paragraph{Pass criterion and metrics.}
Syntax checking is mandatory for all tasks. Quantum-state checking is applied to \emph{classical-logic} and \emph{pulse-control} tasks, while schedule-based checking is applied to \emph{timing scheduling} tasks. For \emph{complex} tasks, both state and schedule checks must pass. We summarize the overall pass condition as:
\[
\mathrm{Pass}(x)=\mathrm{Syn}(x)\wedge\Big(\neg \mathbb{I}_{\mathrm{state}}(x)\,\vee\,\mathrm{State}(x)\Big)\wedge\Big(\neg \mathbb{I}_{\mathrm{sched}}(x)\,\vee\,\mathrm{Sched}(x)\Big),
\]
where $\mathrm{Syn}(x)$ denotes successful parsing/structural validation, $\mathrm{State}(x)$ denotes passing the quantum-state check, and $\mathrm{Sched}(x)$ denotes passing the schedule check. The indicator functions $\mathbb{I}_{\mathrm{state}}(x)$ and $\mathbb{I}_{\mathrm{sched}}(x)$ specify whether a task instance requires state and schedule verification, respectively (e.g., both are true for complex tasks).

We report performance using the standard pass@$k$ metric. Given $n$ generated samples for a task and $c$ correct ones among them, we use the unbiased estimator:
\[
\mathrm{pass@}k=
1-\dfrac{\binom{n-c}{k}}{\binom{n}{k}}
\]
\paragraph{Evaluation Quality}
We assess the fidelity of our simulation and validation pipeline by comparing its judgments against expert annotations. We randomly sampled 200 tasks from the test dataset, collected the answers from GPT-5.6-Terra, and labeled each solution as correct/incorrect using (i) our automated verifier (\emph{Verifier}) and (ii) quantum-programming expert (\emph{Expert}). As shown in Table~\ref{tab:verifier-agreement}, the two agree on 189 out of 200 tasks. For the 11 disagreements, the primary source is ambiguity in interpreting prompt requirements in the \texttt{TODO} description (e.g., whether ``sequence'' refers to a loop or an explicitly unrolled sequence; details in Appendix~\ref{appendix:verifier-fid}). Overall, the agreement rate is 94.5\% and Cohen's $\kappa$ is 0.89, with a 95\% confidence interval of 90.75\%--97\%, indicating high agreement in this audit. Detailed validation of verifier and its newly supported semantics can be found in Appendix~\ref{appendix:verifier-fid}

\begin{table}[!htbp]
\centering
\caption{Agreement between the automated verifier (\emph{Verifier}) and human experts (\emph{Expert})}
\label{tab:verifier-agreement}
\small

\begin{tabular}{
|>{\raggedright\arraybackslash}p{1.8cm}
|>{\centering\arraybackslash}p{1.8cm}
|>{\centering\arraybackslash}p{1.8cm}
|>{\centering\arraybackslash}p{1.8cm}
|>{\centering\arraybackslash}p{1.6cm}
|>{\centering\arraybackslash}p{2.4cm}|
}
\hline
&
\textbf{expert pass} &
\textbf{expert fail} &
\textbf{agreement} &
\textbf{kappa} &
\textbf{95\% CI} \\
\hline

\textbf{verifier pass} &
101 &
7 &
\multirow{2}{*}{94.5\%} &
\multirow{2}{*}{0.89} &
\multirow{2}{*}{90.75\%\textasciitilde{}97\%} \\
\cline{1-3}

\textbf{verifier fail} &
4 &
88 &
&
& \\
\hline
\end{tabular}

\normalsize
\vspace{-0.5em}
\end{table}

\vspace{-1em}
\section{Experiments}
\subsection{Experimental Setup}
\paragraph{Models and Methods}
(1) We evaluate four base models: GPT-5.6-Terra-Medium-Thinking (\texttt{GPT-5.6-Terra}), DeepSeek-V4-Pro (\texttt{DPSK-V4-Pro}), Llama-3.3-70B-Instruct (\texttt{Llama-70B-base}), and Meta-Llama-3.1-8B-Instruct (\texttt{Llama-8B-base}). Additional models are reported in Appendix~\ref{appendix:extraresults}. (2) We evaluate \texttt{GPT-5.6-Terra-Agent}, which combines GPT-5.6-Terra with OpenCode and a RAG (Retrieval-Augmented-Generation) tool over the official OpenQASM documentation.(3) We evaluate cross-theme three-shot prompting~\citep{brown2020fewshot}. Each test problem is paired with three prompt--solution exemplars selected from training-side themes, with the held-out test theme excluded; the corresponding models are denoted by the suffix \texttt{-fs}. (4) We fine-tune Llama-70B and Llama-8B with LoRA on QCircuitBench~\citep{yang2024qcircuitbench}, Agent-Q~\citep{jern2025agentQ}, and the QASM-Eval training split, denoted by the suffixes \texttt{-QCB}, \texttt{-AGQ}, and \texttt{-ours}, respectively. For every model--task pair, we sample five independent solutions and compute pass@$k$ using the standard unbiased estimator.

\paragraph{Training and Testing Details}
We generate 4,000 training tasks, each containing an average of 191 lines and 2,212 tokens, for approximately 8.8M tokens in total. For a controlled comparison, the QCircuitBench and Agent-Q fine-tuning sets are matched to the same token scale. We train each LoRA model for three epochs on two 80GB H100 GPUs, using a context length of 8,192, batch size 8, learning rate $1\times 10^{-5}$, LoRA rank 8, and LoRA $\alpha=8$.

The test set contains 1,200 tasks (100 themes). We partition the themes into five disjoint folds of 20 themes (240 tasks) each. For every fold, the corresponding fine-tuned model is trained without examples from the held-out themes and evaluated only on that fold; aggregating the five folds therefore covers all 1,200 test tasks while preserving strict theme-level separation. 

\subsection{Experimental Results}
Table~\ref{tab:main-results} summarizes overall performance under the theme-held-out protocol. \textbf{The base-model results establish substantial headroom for adaptation}: GPT-5.6-Terra reaches 39.65\% pass@1, followed by DPSK-V4-Pro at 25.67\%, Llama-70B-base at 15.70\%, and Llama-8B-base at 9.98\%. Their pass@5 results show that additional sampling provides useful gains for every model while maintaining the same ordering.

\textbf{Cross-theme exemplars transfer effectively to unseen task families.} Three-shot prompting raises pass@1 to 63.07\% for GPT-5.6-Terra-fs, 60.10\% for DPSK-V4-Pro-fs, 55.28\% for Llama-70B-fs, and 32.87\% for Llama-8B-fs. Because no exemplar shares the test theme, these gains reflect the models' ability to infer reusable correspondences between natural-language constraints and OpenQASM~3 implementations.

\textbf{QASM-Eval fine-tuning provides the strongest Llama-family results on held-out themes.} Llama-8B-ours reaches 46.93\% pass@1 and 60.25\% pass@5, while Llama-70B-ours reaches 61.17\% pass@1 and 72.92\% pass@5. The corresponding syntax-valid rates of 90.38\% and 92.57\% show that the learned representation transfers beyond the training themes and supports reliable generation at both model scales.

\begin{table}[!htbp]
\centering
\captionsetup{skip=3pt}
\caption{Overall results on the 1,200-task theme-held-out QASM-Eval test set. Bracketed values are 95\% theme-cluster bootstrap confidence intervals. ``-Agent'' denotes coding agent equipped with RAG tool; ``-fs'' denotes cross-theme three-shot prompting; ``-QCB'', ``-AGQ'', and ``-ours'' denote LoRA fine-tuning on QCircuitBench~\citep{yang2024qcircuitbench}, Agent-Q~\citep{jern2025agentQ}, and the QASM-Eval training set, respectively.}
\label{tab:main-results}
\small
\resizebox{\linewidth}{!}{%
\begin{tabular}{lccccc}
\toprule
Model & pass@1 & pass@1 95\% CI & pass@5 & pass@5 95\% CI & Syntax-valid rate \\
\midrule
GPT-5.6-Terra      & 39.65\% & [35.67\%, 43.58\%] & 49.50\% & [45.42\%, 53.50\%] & 67.03\% \\
DPSK-V4-Pro    & 25.67\% & [21.68\%, 29.80\%] & 38.25\% & [33.92\%, 42.58\%] & 54.03\% \\
Llama-70B-base     & 15.70\% & [12.05\%, 19.62\%] & 18.83\% & [14.67\%, 23.17\%] & 32.77\% \\
Llama-8B-base      &  9.98\% & [ 7.58\%, 12.55\%] & 18.67\% & [14.92\%, 22.58\%] & 40.50\% \\
\midrule
GPT-5.6-Terra-Agent & 47.37\% & [43.90\%, 50.73\%] & 55.25\% & [51.58\%, 58.75\%] & 75.72\% \\
\midrule
GPT-5.6-Terra-fs   & 63.07\% & [59.58\%, 66.42\%] & 71.17\% & [67.25\%, 74.92\%] & 93.57\% \\
DPSK-V4-Pro-fs & 60.10\% & [56.70\%, 63.38\%] & 71.58\% & [67.75\%, 75.25\%] & 93.47\% \\
Llama-70B-fs       & 55.28\% & [51.58\%, 58.85\%] & 69.42\% & [65.83\%, 72.92\%] & 81.82\% \\
Llama-8B-fs        & 32.87\% & [30.02\%, 35.65\%] & 53.25\% & [49.58\%, 56.83\%] & 63.85\% \\
\midrule
Llama-70B-QCB      & 17.33\% & [15.98\%, 18.70\%] & 49.50\% & [46.75\%, 52.25\%] & 39.53\% \\
Llama-8B-QCB       &  8.67\% & [ 7.80\%,  9.53\%] & 30.67\% & [28.08\%, 33.17\%] & 28.52\% \\
Llama-70B-AGQ      & 16.05\% & [14.70\%, 17.38\%] & 46.75\% & [44.00\%, 49.50\%] & 35.53\% \\
Llama-8B-AGQ       &  5.83\% & [ 5.12\%,  6.58\%] & 23.42\% & [21.00\%, 25.92\%] & 23.20\% \\
\midrule
Llama-70B-ours     & 61.17\% & [57.40\%, 64.78\%] & 72.92\% & [69.08\%, 76.67\%] & 92.57\% \\
Llama-8B-ours      & 46.93\% & [42.73\%, 51.05\%] & 60.25\% & [55.25\%, 65.08\%] & 90.38\% \\
\bottomrule
\end{tabular}%
}
\vspace{-0.45em}
\end{table}

\subsection{Performance Analysis}
\begin{figure}[!t]
  \centering
  \captionsetup{skip=3pt}
  \begin{minipage}[t]{0.49\linewidth}
    \centering
    \includegraphics[width=\linewidth,trim=0 24bp 0 4bp,clip]{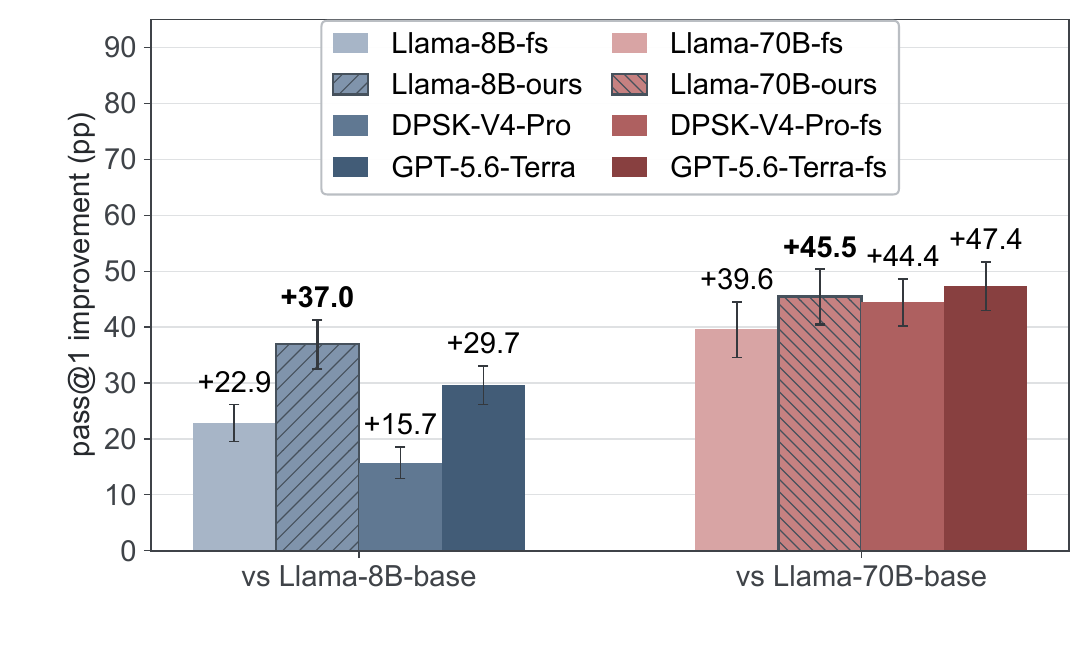}
    \caption{Paired pass@1 improvements relative to the corresponding Llama base model. Error bars show paired theme-cluster bootstrap 95\% confidence intervals.}
    \label{fig:paired-improvements}
  \end{minipage}\hfill
  \begin{minipage}[t]{0.49\linewidth}
    \centering
    \includegraphics[width=\linewidth,trim=0 24bp 0 4bp,clip]{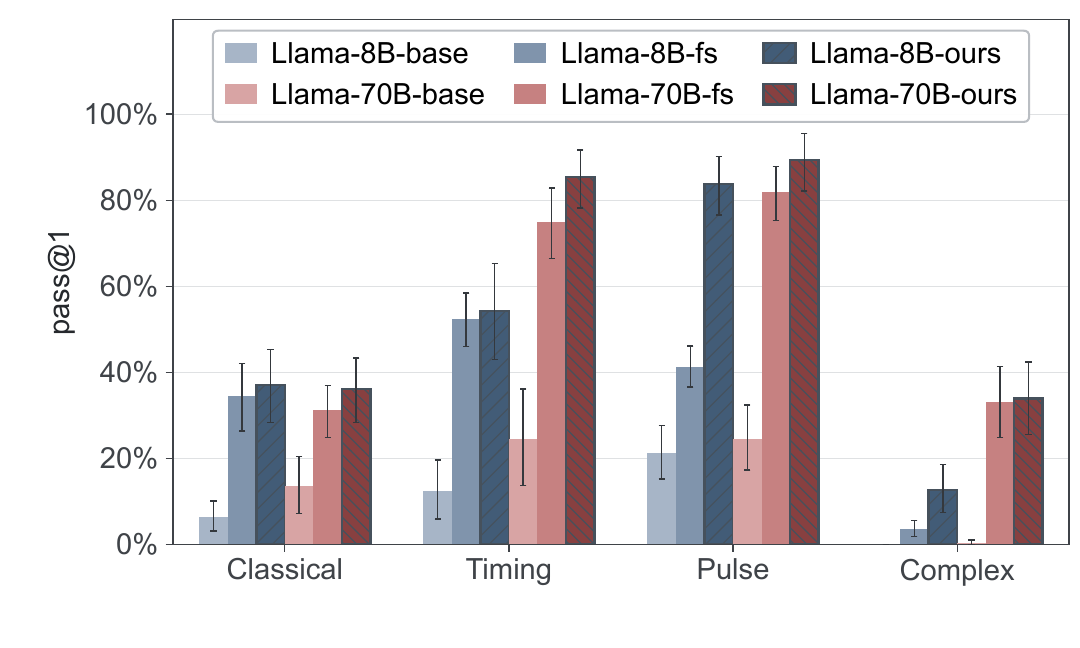}
    \caption{Domain-level pass@1 for the Llama base, cross-theme three-shot, and QASM-Eval fine-tuned models. Error bars show theme-cluster bootstrap 95\% confidence intervals.}
    \label{fig:domain-pass1}
  \end{minipage}
  \vspace{-0.45em}
\end{figure}

\paragraph{Held-out-theme improvements and domain transfer}
\textbf{Figure~\ref{fig:paired-improvements}: Adaptation produces large and statistically reliable gains on unseen themes.} For Llama-8B, cross-theme three-shot prompting improves pass@1 by 22.88 percentage points (95\% CI: 19.52--26.18), whereas QASM-Eval fine-tuning improves it by 36.95 points (32.53--41.27). For Llama-70B, the corresponding gains are 39.58 points (34.55--44.45) and 45.47 points (40.45--50.33). All eight displayed contrasts have paired intervals excluding zero and remain significant after Holm correction ($p_{\mathrm{Holm}}=8\times10^{-5}$), supporting a consistent adaptation effect at the theme level.

\textbf{Figure~\ref{fig:domain-pass1}: Fine-tuning transfers across all four OpenQASM~3 domains.} For Llama-8B, pass@1 rises from 6.3\% to 37.1\% on classical tasks, from 12.4\% to 54.2\% on timing tasks, from 21.2\% to 83.8\% on pulse tasks, and from 0.0\% to 12.7\% on complex tasks. Llama-70B-ours reaches 85.3\% on timing, 89.3\% on pulse, and 33.9\% on complex tasks. The timing and pulse gains are especially informative because they require the generated program to realize physical scheduling and pulse-control behavior, while the complex-domain gain shows progress in composing multiple OpenQASM~3 capabilities.

\begin{figure}[!t]
  \centering
  \captionsetup{skip=3pt}
  \begin{minipage}[t]{0.49\linewidth}
    \centering
    \includegraphics[width=\linewidth,trim=0 8bp 0 4bp,clip]{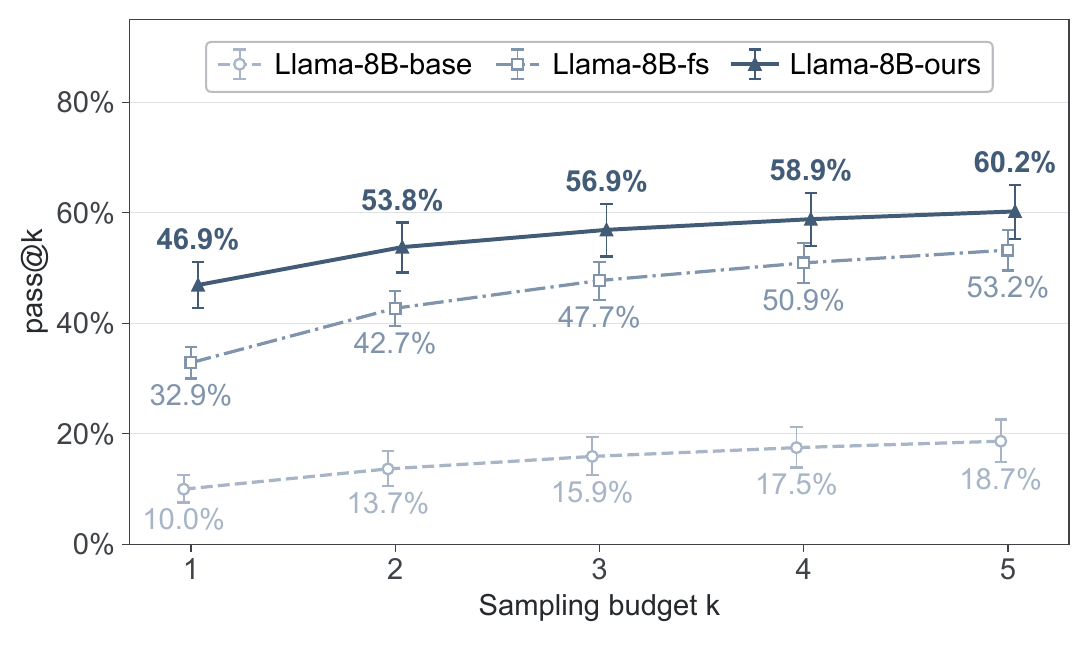}
    \centerline{\small (a) Llama-8B}
  \end{minipage}\hfill
  \begin{minipage}[t]{0.49\linewidth}
    \centering
    \includegraphics[width=\linewidth,trim=0 8bp 0 4bp,clip]{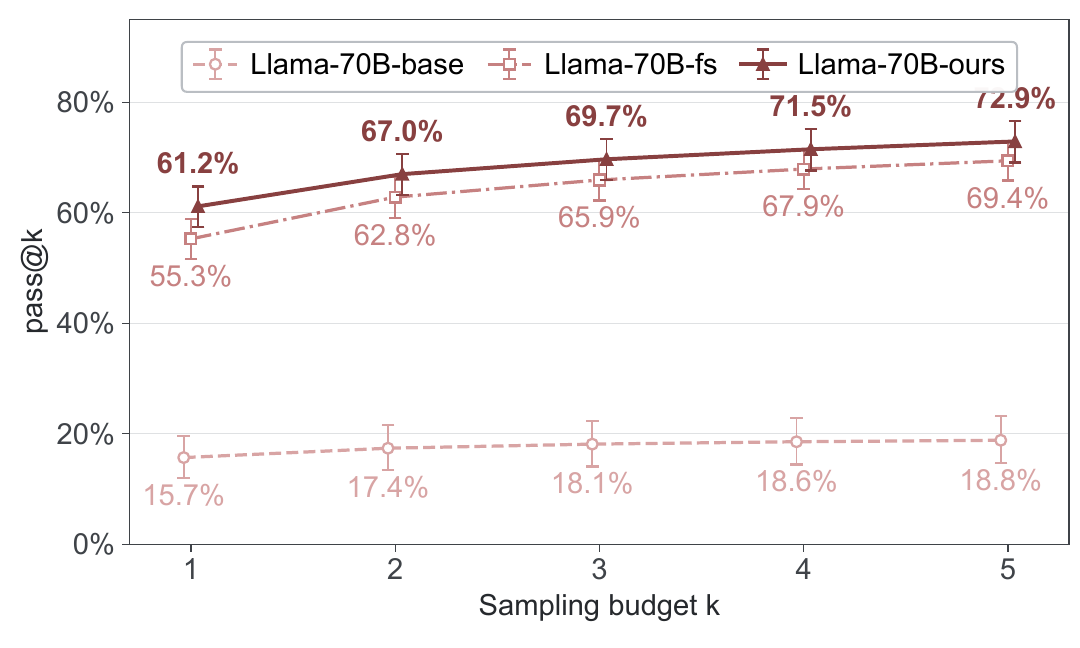}
    \centerline{\small (b) Llama-70B}
  \end{minipage}
  \caption{Unbiased pass@$k$ estimates for (a) Llama-8B and (b) Llama-70B under base, cross-theme three-shot, and QASM-Eval fine-tuned settings. Error bars show theme-cluster bootstrap 95\% confidence intervals.}
  \label{fig:passk}
  \vspace{-0.45em}
\end{figure}

\paragraph{Effect of the sampling budget}
\textbf{Figure~\ref{fig:passk}: Fine-tuning improves first-attempt reliability, while additional samples provide complementary gains.} For Llama-8B, pass@1/pass@5 increases from 9.98\%/18.67\% for the base model to 32.87\%/53.25\% with cross-theme prompting and 46.93\%/60.25\% after fine-tuning. For Llama-70B, the corresponding values are 15.70\%/18.83\%, 55.28\%/69.42\%, and 61.17\%/72.92\%. The fine-tuned models lead at every $k$, while the upward curves show that sampling remains a useful complement once the probability of a correct individual solution has been raised.

\begin{figure}[!t]
  \centering
  \captionsetup{skip=3pt}
  \begin{minipage}[t]{0.49\linewidth}
    \centering
    \includegraphics[width=\linewidth,trim=0 24bp 0 4bp,clip]{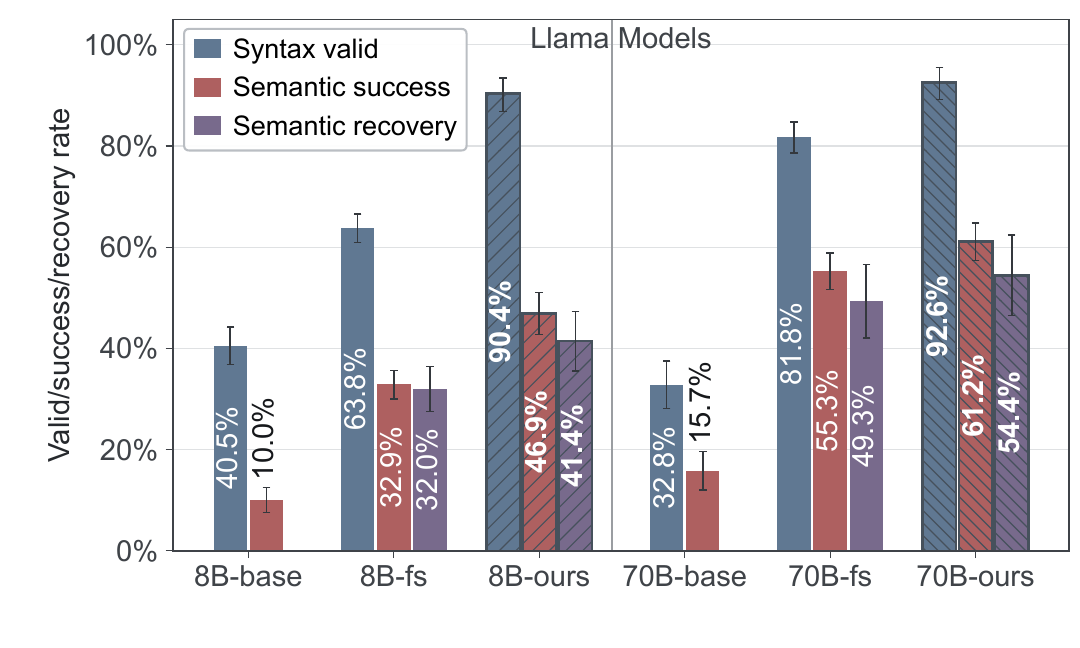}
    \centerline{\small (a) Syntax validity and semantic competence}
  \end{minipage}\hfill
  \begin{minipage}[t]{0.49\linewidth}
    \centering
    \includegraphics[width=\linewidth,trim=0 24bp 0 4bp,clip]{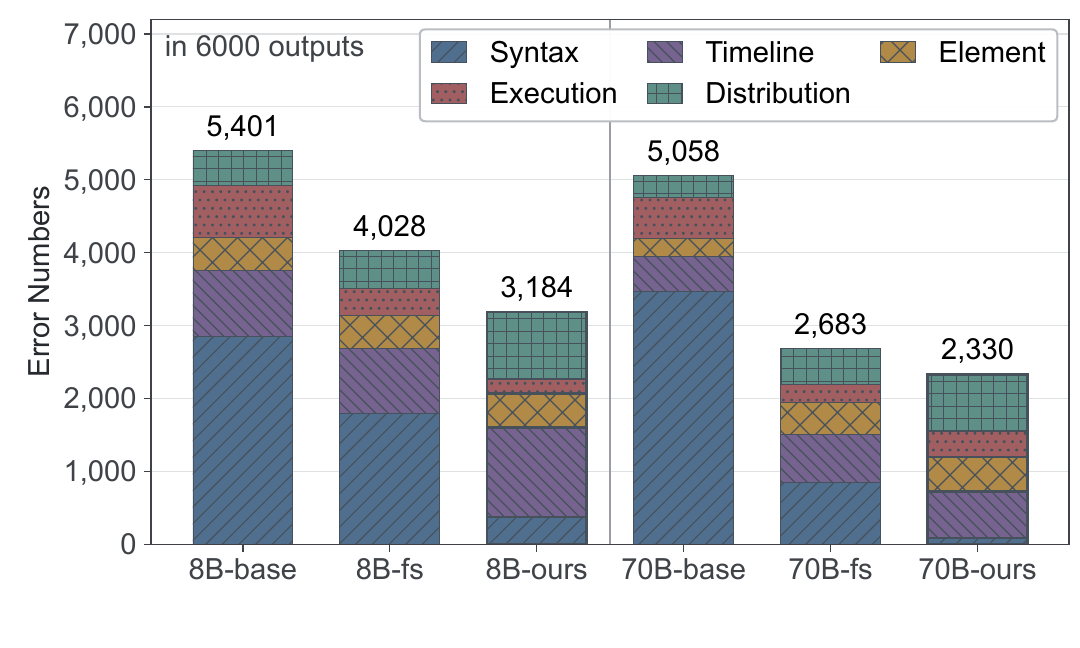}
    \centerline{\small (b) Absolute outcome and error counts}
  \end{minipage}
  \caption{Analysis of syntactic and semantic competence. (a) Syntax validity, complete semantic success, and semantic recovery for the Llama models (b) Error counts among 6,000 generated solutions per model.}
  \label{fig:semantic-error-analysis}
  \vspace{-0.45em}
\end{figure}

\paragraph{Syntax, quantum semantics, and residual errors}
\textbf{Figure~\ref{fig:semantic-error-analysis}(a): Fine-tuning improves task-semantic success in addition to OpenQASM syntax.} For Llama-8B, syntax validity increases from 40.5\% to 90.4\%, while complete semantic success increases from 10.0\% to 46.9\%; for Llama-70B, the corresponding increases are 32.8\% to 92.6\% and 15.7\% to 61.2\%. Moreover, among tasks where the base model already produced syntax-valid but semantically incorrect outputs, semantic recovery reaches 41.4\% for Llama-8B-ours versus 32.0\% for Llama-8B-fs, and 54.4\% for Llama-70B-ours versus 49.3\% for Llama-70B-fs. These recovery gains show improved translation of quantum tasks into OpenQASM~3 behavior under the benchmark verifier, beyond surface-level syntax repair.

\textbf{Figure~\ref{fig:semantic-error-analysis}(b): Adaptation substantially reduces the absolute failure burden at both model scales.} Across 6,000 outputs, the number of unsuccessful generations decreases from 5,401 to 3,184 for Llama-8B and from 5,058 to 2,330 for Llama-70B; syntax errors decrease from 2,858 to 378 and from 3,474 to 87, respectively. As more programs become executable, the residual error mix naturally shifts toward timeline, distribution, execution, and required-element checks, which provide a more informative view of the remaining quantum-semantic constraints. Together with the simultaneous rise in complete semantic success in Figure~\ref{fig:semantic-error-analysis}(a), this pattern shows that fine-tuning advances both syntactic reliability and end-to-end OpenQASM~3 competence.

\section{Conclusion}
QASM-Eval is the first dataset for training and evaluating LLMs on hardware-facing OpenQASM~3 features, spanning classical logic, explicit timing, and pulse control. Its simulation and scheduling verifier assesses syntax and task-specified behavior under a common execution model. Under theme-held-out evaluation, targeted fine-tuning produces substantial gains and enables mid-scale models to approach few-shot GPT-5.6-Terra. By releasing the dataset, generation pipeline, fine-tuned models, and verifier, we aim to support reliable LLM assistants for translating specifications into hardware-facing quantum programs.

\clearpage
\setlength{\textfloatsep}{\originaltextfloatsep}
\setlength{\floatsep}{\originalfloatsep}
\bibliographystyle{iclr2027_conference}
\bibliography{refs}

\clearpage
\appendix
\raggedbottom
\setlength{\floatsep}{8pt plus 2pt minus 2pt}
\setlength{\textfloatsep}{9pt plus 2pt minus 2pt}
\setlength{\intextsep}{8pt plus 2pt minus 2pt}
\renewcommand{\topfraction}{0.95}
\renewcommand{\bottomfraction}{0.95}
\renewcommand{\textfraction}{0.03}
\renewcommand{\floatpagefraction}{0.80}
\setcounter{topnumber}{5}
\setcounter{bottomnumber}{5}
\setcounter{totalnumber}{8}
\captionsetup{font=small,skip=4pt}

\setlength{\parskip}{3.5pt plus 1pt minus 0.5pt}
\section{Dataset Details}
\begingroup\small
\label{appendix:datasetdetails}

This section provides additional details of the dataset, including the structure of the background program and the full set of themes used in the core tasks.

The background program defines the objects and reference settings required by the downstream core tasks. These include the drive and measurement frequencies of each qubit, the definitions of pulse interfaces (ports), waveform specifications such as Gaussian and DRAG pulses, classical function definitions, the initial frames of waveforms, and randomly instantiated pulse operations and gate sequences. Based on this background, the pipeline randomly generates core tasks according to predefined templates.

We further enumerate all themes covered by the core tasks from Table \ref{tab:classical-themes} to \ref{tab:core-task-themes}. The corresponding syntax categories and language features have already been summarized in Table~\ref{tab:task-categories}. Within each theme, different variants may alter the code structure, statement order, operated objects, parameter values, and logical decision criteria. However, these variations remain confined to the scope of the given theme, ensuring that each task family consistently targets a specific syntax pattern or language feature.

\begin{minipage}[t]{.485\linewidth}\centering\captionsetup{type=table,font=footnotesize}
\fontsize{7.2}{8}\selectfont
\renewcommand{\arraystretch}{0.70}
\centering
\caption{Classical Logic themes}
\label{tab:classical-themes}
\begin{tabular}{cl}
\toprule
ID & Theme \\
\midrule
1  & measure\_then\_reset\_branch \\
2  & repeat\_until\_success \\
3  & conditional\_correction\_multibit \\
4  & for\_loop\_continue\_pattern \\
5  & while\_loop\_break\_on\_measure \\
6  & angle\_arithmetic\_mod\_wrap \\
7  & angle\_scaled\_by\_uint \\
8  & float\_expr\_cast\_to\_angle \\
9  & integer\_arithmetic\_controls\_flow \\
10 & mixed\_comparison\_with\_cast \\
11 & bit\_shift\_and\_mask\_test \\
12 & popcount\_triggered\_gate \\
13 & rotl\_rotr\_pattern\_match \\
14 & bitwise\_and\_xor\_pipeline \\
15 & integer\_switch\_on\_computed\_index \\
16 & switch\_on\_int\_from\_bit\_literal \\
17 & set\_iteration\_discrete \\
18 & array\_iteration\_angles\_or\_ints \\
19 & single\_extern\_call \\
20 & nested\_control\_switch\_in\_loop\_with\_if \\
21 & measurement\_accumulation\_counter \\
22 & early\_termination\_end \\
23 & bit\_slice\_alias\_and\_iteration \\
24 & membership\_test\_in\_set \\
25 & switch\_const\_expression\_cases\_no\_default \\
\bottomrule
\end{tabular}
\end{minipage}\hfill\begin{minipage}[t]{.485\linewidth}\centering\captionsetup{type=table,font=footnotesize}
\fontsize{7.2}{8}\selectfont
\renewcommand{\arraystretch}{0.70}
\centering
\caption{Timing and scheduling themes}
\label{tab:Timing Scheduling themes}
\begin{tabular}{cl}
\toprule
ID & Theme \\
\midrule
1  & basic\_delay\_units \\
2  & dt\_delay \\
3  & duration\_arithmetic \\
4  & mixed\_units\_duration \\
5  & left\_alignment\_stretch \\
6  & right\_alignment\_stretch \\
7  & center\_alignment\_stretch \\
8  & proportional\_placement \\
9  & durationof\_single\_gate \\
10 & durationof\_subcircuit \\
11 & dynamic\_padding\_safe \\
12 & basic\_box\_boundary \\
13 & timed\_box\_known\_duration \\
14 & box\_with\_stretch\_fill \\
15 & barrier\_ordering \\
16 & hahn\_echo\_midpoint \\
17 & center\_align\_halfdiff\_known \\
18 & multi\_qubit\_delay\_semantics \\
19 & nop\_sync\_in\_box \\
20 & duration\_update \\
21 & explicit\_delay\_as\_structure \\
22 & delay\_zero\_ordering \\
23 & nested\_box \\
24 & multi\_stretch\_solve\_system \\
25 & durationof\_on\_box\_compound \\
\bottomrule
\end{tabular}
\end{minipage}

\medskip
\begin{minipage}[t]{.485\linewidth}\centering\captionsetup{type=table,font=footnotesize}
\fontsize{7.2}{8}\selectfont
\renewcommand{\arraystretch}{0.70}
\centering
\caption{Pulse Control themes}
\label{tab:pulse-themes}
\begin{tabular}{cl}
\toprule
ID & Theme \\
\midrule
1  & x\_gaussian\_play \\
2  & virtual\_z\_shift\_phase \\
3  & measure\_capture\_bit \\
4  & sx\_drag\_play \\
5  & cr\_composite\_multi\_frame \\
6  & active\_reset\_equal\_time\_branches \\
7  & sideband\_modulation\_mix \\
8  & raw\_samples\_waveform\_literal \\
9  & frame\_sync\_barrier \\
10 & raw\_capture\_trace \\
11 & defcal\_q\_vs\_physical\_qubit \\
12 & multiplexed\_readout\_multi\_frame \\
13 & simultaneous\_plays\_parallel\_frames \\
14 & global\_cal\_scope\_reuse \\
15 & phase\_tracking\_with\_time\_advance \\
16 & dd\_sequence\_delay\_play \\
17 & waveform\_dsp\_add\_scale\_phase\_shift \\
18 & frequency\_control\_set\_shift \\
19 & frame\_state\_get\_set\_swap \\
20 & defcal\_matching\_priority \\
21 & newframe\_time\_origin\_cal\_vs\_defcal \\
22 & compile\_time\_determinable\_duration \\
23 & frame\_collision\_avoidance \\
24 & multi\_frames\_same\_port\_patterns \\
25 & measurement\_return\_type\_variants \\
\bottomrule
\end{tabular}
\end{minipage}\hfill\begin{minipage}[t]{.485\linewidth}\centering\captionsetup{type=table,font=footnotesize}
\fontsize{7.2}{8}\selectfont
\renewcommand{\arraystretch}{0.70}
\centering
\caption{Complex Tasks themes}
\label{tab:core-task-themes}
\begin{tabular}{cl}
\toprule
ID & Theme \\
\midrule
1  & active\_reset\_loop \\
2  & ramsey\_feedback\_phase\_comp \\
3  & hahn\_echo\_characterization \\
4  & t1\_relaxation\_gated\_readout \\
5  & rabi\_amplitude\_scan \\
6  & qubit\_spectroscopy\_scan \\
7  & echoed\_cr\_gate\_verification \\
8  & dynamic\_cnot\_spectator\_comp \\
9  & multiplexed\_readout \\
10 & raw\_waveform\_capture\_and\_filter \\
11 & measurement\_crosstalk\_calibration \\
12 & realtime\_feedback\_correction \\
13 & pipeline\_measure\_reset\_prep \\
14 & randomized\_benchmarking\_controller \\
15 & repeat\_until\_success\_prep \\
16 & leakage\_detection\_and\_recovery \\
17 & virtual\_z\_phase\_tracking\_test \\
18 & calibration\_hot\_swap\_local\_override \\
19 & durationof\_alignment\_scheduling \\
20 & boxed\_dynamic\_decoupling \\
21 & switch\_routed\_feedforward \\
22 & in\_shot\_micro\_averaging \\
23 & late\_as\_possible\_conditional \\
24 & timeout\_active\_reset\_with\_update \\
25 & syndrome\_feedforward\_idle\_scheduling \\
\bottomrule
\end{tabular}
\end{minipage}
\FloatBarrier

\endgroup

\subsection{Quantum Programming Basics}
\label{appendix:quantum-basics}
Before explaining each theme in detail, we first introduce the quantum-programming fundamentals covered by the dataset and explain the significance of these advanced operations for quantum experiments. Quantum programs coordinate a quantum state, classical measurement records, and physical control resources. A gate transforms the state coherently; a measurement produces an outcome and changes the conditional state; a controller uses that outcome to select subsequent operations. OpenQASM~3 brings these layers together through classical control flow, explicit timing, and calibration blocks~\citep{cross2022openqasm3}. The following concepts explain the physical dependencies exercised by the four task categories.

\paragraph{Mid-circuit measurement and quantum error correction}
Because measurement changes the quantum state, branches conditioned on mid-circuit measurements are more complex in quantum programs than in classical programs. For an ideal computational-basis measurement of a qubit, the projectors are $P_m=|m\rangle\langle m|$, with $m\in\{0,1\}$. Given a density operator $\rho$, outcome $m$ occurs with probability $p_m=\operatorname{Tr}(P_m\rho)$ and leaves the conditional state $\rho_m=P_m\rho P_m/p_m$. When the measured qubit belongs to a larger register, the projector acts jointly with the identity on the unmeasured subsystem. Consequently, measuring one qubit can condition the state of other, entangled qubits. If the controller applies a subsequent operation $U_m$, the resulting ensemble is
\begin{equation}
\mathcal{E}(\rho)=\sum_m U_m P_m\rho P_m U_m^\dagger.
\label{eq:app-measure-feedback}
\end{equation}
This equation explains why a measurement outcome has two roles: it records which branch occurred and identifies the state on which the selected operation acts. Mid-circuit measurement makes this information available while the rest of the quantum computation continues. Applications include active reset, conditional state preparation, and repeated error-syndrome extraction~\citep{ryan2021RealQEC}.

Mid-circuit measurement is especially important for quantum error correction, which encodes logical information across several physical qubits. In the stabilizer framework, an encoded state lies in the joint $+1$ eigenspace of commuting Pauli operators $S_j$. An error can move the state into a different eigenspace, and a syndrome measurement obtains the corresponding eigenvalues $s_j\in\{+1,-1\}$. These values describe error information while preserving the encoded logical amplitudes within the measured subspace~\citep{gottesman2010QEC}. For a two-qubit parity check, for example,
\begin{equation}
P_\pm=\frac{I\pm Z_1Z_2}{2},\qquad
P_+\!\left(\alpha|00\rangle+\beta|11\rangle\right)
=\alpha|00\rangle+\beta|11\rangle.
\end{equation}
An ancilla can acquire this parity through entangling gates and then be measured. The controller associates the syndrome with a correction or an update to its record of pending Pauli corrections. Repeated rounds require consistent ancilla roles, measurement order, and feedback latency. Real-time fault-tolerant error correction therefore joins quantum-state evolution with classical processing during the circuit~\citep{ryan2021RealQEC}. Syndrome-inspired tasks in our complex category exercise components of this interaction: entanglement, readout, conditional correction, and scheduling of other qubits.

\paragraph{Timing controls and dynamical decoupling}
  Quantum computing requires precise control on operation timings. A qubit evolves during an idle interval. Under a simple detuning Hamiltonian $H=\hbar\delta Z/2$, waiting for time $t$ produces $U(t)=\exp(-i\delta t Z/2)$, so relative phase depends on the duration. Energy relaxation and fluctuations in the environment also make the time spent in a state relevant to its later use~\citep{krantz2019coherencetime}. Gates, readout, and feedback must therefore be positioned with respect to one another. A fixed-duration box establishes a control window; a barrier aligns participating resources; an explicit delay advances a selected resource; and a stretch variable allocates available slack under the stated constraints~\citep{cross2022openqasm3}. These operations determine the physical intervals over which the system evolves.

Timing control is the basis of Dynamical decoupling. Dynamical decoupling is a technique that suppresses environmental noise in quantum systems through precisely applied pulses. For ideal instantaneous $\pi$ pulses and dephasing noise $\delta(t)$, the accumulated phase can be written
\begin{equation}
\Phi(T)=\int_0^T y(t)\delta(t)\,dt,\qquad y(t)\in\{+1,-1\},
\label{eq:app-dd-phase}
\end{equation}
where the switching function changes sign at each refocusing pulse. In a Hahn echo with quasi-static detuning, equal free-evolution intervals on the two sides of a central $\pi$ pulse cancel the accumulated phase. Longer sequences select a temporal response to noise through their pulse positions and axes~\citep{khodjasteh2005DD,bylander2011DD_timing}. Thus, the placement of slack and the phase of each pulse are critical parts of a decoupling sequence.

\paragraph{Pulses and calibration.}
On common quantum-computing platforms, operations on qubits are physically implemented using control pulses, such as laser or microwave pulses. For example, for a driven two-level system in a rotating frame, a useful model is
\begin{equation}
H(t)/\hbar=\frac{1}{2}\left[\Delta(t)Z+\Omega_x(t)X+\Omega_y(t)Y\right],
\qquad U(T)=\mathcal{T}\exp\!\left[-\frac{i}{\hbar}\int_0^T H(t)\,dt\right].
\label{eq:app-control-hamiltonian}
\end{equation}
The drive envelope sets $\Omega_x$ and $\Omega_y$, while the difference between drive and qubit frequencies contributes to $\Delta$. With a resonant drive of fixed phase $\phi$, the envelope area sets a rotation angle and the phase selects an axis in the equatorial plane. Pulse shape also controls spectral selectivity. For weakly anharmonic qubits, DRAG adds a derivative component in the orthogonal quadrature to suppress leakage into higher levels~\citep{motzoi2009DRAGpulse}; optimal-control methods shape fields to realize target transformations in coupled systems~\citep{khaneja2005DMRpulse}.

Fine-grained pulse control, in turn, enables calibration. Because of intrinsic noise in quantum systems and imperfections in hardware or the environment, maintaining high fidelity on quantum-computing platforms requires regular calibration. In simple terms, calibration uses experiments to characterize and adjust parameters such as drive frequency, amplitude, duration, phase, and pulse shape~\citep{burnett2019drift1,proctor2020drift2}. OpenQASM Pulse (OpenPulse) supports the operations required for calibration, including applying and manipulating waveforms, configuring ports, and acquiring signals.

\subsection{Details of the Tasks}
\label{appendix:theme-meaning}
The themes in Tables~\ref{tab:classical-themes}--\ref{tab:core-task-themes} are grouped below by their quantum role. C, T, P, and X denote the classical, timing, pulse, and complex theme identifiers, respectively. 

\subsubsection{Classical decisions acting on quantum systems}
\paragraph{Measurement-driven state preparation and control (C01--C05, C16, C20--C22).}
These tasks combine measurement with branches, bounded repetition, early termination, and nested decisions. They describe how a controller reacts to the state revealed by measurement: reset an excited qubit, select a subsequent gate, accumulate outcomes, or stop a preparation procedure after a designated result. The ordering of measurement and action determines the branch state in Eq.~\eqref{eq:app-measure-feedback}. Repeated measurements also act on the state left by earlier iterations. A successful completion must carry this evolving state through the loop and preserve the stopping condition that defines the terminal ensemble.

\paragraph{Numerical control of rotations (C06--C10, C15, C17--C19).}
Angle arithmetic, casts, arrays, arithmetic dispatch, and external controller functions supply parameters for quantum operations. The quantum effect is determined by the resulting rotation, $R_\alpha(\theta)=\exp(-i\theta\sigma_\alpha/2)$, including its axis, angle, and position in the sequence. Array iteration may describe a prescribed sequence of rotations or a series of measurement-conditioned actions. Controller updates connect a stored classical quantity to the next physical operation. These themes test whether the generated arithmetic preserves that connection across data representations and repeated execution.

\paragraph{Encoding and routing measurement information (C11--C14, C23--C25).}
Bit masks, shifts, population counts, register slices, and multiway dispatch determine which outcomes influence an operation and which qubits receive it. In quantum feedback, bits can encode different syndrome components or the outcomes of distinct measurements. Their positions carry the identities of those observations. A mask that selects a different bit routes different physical information into the decision; a changed subset of qubits alters the support of the correction. The common requirement across these themes is to retain the association among measured subsystem, classical record, and quantum action.

\subsubsection{Time and resource constraints}
\paragraph{Representing and deriving physical durations (T01--T04, T09--T11, T20, T25).}
The duration themes express waits in physical units and device ticks, combine durations, and derive intervals from calibrated gate or subcircuit durations. Their quantum role is to make the schedule agree with the actual duration of the operation being composed. For a target window $T$ containing a calibrated operation of duration $d$, the available padding is $T-d$. Referencing $d$ through the operation's duration keeps the timing relation tied to the calibration. This is useful when the same control pattern is instantiated with different gates or hardware parameters.

\paragraph{Placing control events inside a window (T05--T08, T12--T14, T17, T19, T23--T24).}
Left, right, centered, and ratio-constrained alignment distribute slack around gates. Nested boxes and coupled stretch variables add dependencies between inner and outer windows. The relevant quantity is the time of each physical event relative to its neighbors. For example, centering an operation requires equal leading and trailing gaps $(T-d)/2$; a two-to-one ratio assigns $2(T-d)/3$ before the operation and $(T-d)/3$ after it. These placements support synchronized readout, balanced control sequences, and execution within a shared deadline. The generated program must solve the joint constraints imposed by the participating resources.

\paragraph{Refocusing and synchronization (T15--T16, T18, T21--T22).}
Barriers, multi-qubit waits, explicit idles, and zero-duration ordering specify which resources advance together and which operations occur first. Hahn-echo timing places the refocusing operation at the point needed for phase cancellation. Equation~\eqref{eq:app-dd-phase} shows the physical reason: changing a pulse's position changes the intervals with opposite signs in the phase integral. A synchronization constraint also matters when one qubit is read out while another retains quantum information. These themes make local event times and shared resource boundaries explicit.

\subsubsection{Pulse resources and persistent frame state}
\paragraph{Shaping the control field (P01, P04--P05, P07--P08, P16--P17).}
Gaussian and DRAG envelopes, modulated signals, explicit complex samples, and waveform transformations construct the fields that act on qubits. Multi-frame and decoupling sequences combine those envelopes over time. Amplitude scaling changes drive strength, complex phase changes a rotation axis, and envelope duration changes both pulse area and spectral content. Composing these operations requires the model to connect the waveform representation to the control Hamiltonian in Eq.~\eqref{eq:app-control-hamiltonian}. In particular, the meaning of an envelope depends on the frame on which it is played.

\paragraph{Maintaining phase, frequency, and calibration context (P02, P11, P14--P15, P18--P21).}
These themes manipulate virtual phase updates, accumulated frame state, frequency detuning and restoration, and gate calibration selection. A reused frame carries its existing state into the next block. Relative phase updates compose additively, and frequency changes alter the carrier used by subsequent pulses. Gate-specific calibrations bind these controls to an operation and its physical qubits. The tasks therefore exercise contextual reasoning: the effect of a local instruction depends on the phase, frequency, time, and calibration already established by the surrounding program.

\paragraph{Acquisition and resource scheduling (P03, P06, P09--P10, P12--P13, P22--P25).}
Readout tasks combine measurement envelopes with capture operations that return either a classified bit or a raw signal. Their consumers differ: a bit can drive feedback directly, while a raw trace supports further signal processing. Equal-time branches, barriers, parallel frames, fixed-duration blocks, and shared-port cases determine when these operations occur and which resources they occupy. The physical workflow couples stimulus and acquisition. A completion must preserve both the signal being measured and the timing relationships that make its result available to the controller.

\subsubsection{Composed quantum-control workflows}
\paragraph{Characterization and coherent control (X02--X08, X14, X17--X20).}
Ramsey, Hahn-echo, relaxation, Rabi, and frequency-scan themes combine prepared states, pulses, waiting periods, and observations. Each sequence probes a different aspect of the dynamics: accumulated relative phase, refocusing, population decay, drive strength, or resonance response. Echoed interactions and spectator control add relationships between qubits, while phase tracking and local frequency overrides connect successive pulse blocks. A characterization parameter acquires its meaning through its location in the experiment. Moving a delay past an analysis pulse, for example, changes which part of the evolution is converted into a measured population. Randomized-benchmarking-inspired sequences similarly depend on the ordered composition of gates~\citep{knill2008randomized}.

\paragraph{Acquisition, preparation, and bounded feedback (X01, X09--X13, X15--X16, X21--X24).}
These workflows combine readout windows, multiple acquisition paths, conditional preparation, outcome routing, and bounded retry loops. The controller acts while quantum state and frame context persist. Measurement results determine the next intervention, the number of attempts, and the time at which the state becomes available. A loop that terminates on success selects a conditional ensemble, while a loop that aggregates several readouts retains the backaction of earlier measurements. Multi-qubit pipelines additionally require the model to distinguish the qubit being observed from the one being driven or held idle.

\paragraph{Syndrome extraction with scheduled idles (X25).}
The syndrome family brings entangling gates, ancilla readout, feedback, and data-qubit waiting into one control sequence. Its significance lies in their joint causal structure: an ancilla acquires information about a data subsystem, the measurement exposes that information to the controller, and a correction acts on the designated data qubit. Scheduling must also account for the other qubits during readout and feedback. These are the constituent dependencies of repeated syndrome processing discussed in Appendix~\ref{appendix:quantum-basics}. The complex category tests whether a generated completion preserves such dependencies when several individually familiar operations appear together.

\section{Dataset Examples}
\label{appendix:datasetexamples}

In this section, we present a concrete dataset instance to illustrate the structure of a full example. Each instance consists of a background program, a core task, a natural-language TODO prompt that describes the core task, and several additional variants derived from the same task family. The background program is shown in two parts: the declarations, constants, calibration resources, and frame definitions appear in Figure~\ref{fig:dataset-example-background-part1}, while the classical initialization and subsequent gate-level and timing context appear in Figure~\ref{fig:dataset-example-background-part2}. A representative core task from the classical-logic category, specifically the \texttt{nested\_control\_switch\_in\_loop\_with\_if} theme, is shown in Figure~\ref{fig:dataset-example-core-task}. Its corresponding TODO prompt is shown in Figure~\ref{fig:dataset-example-todo}. Three additional variants from the same theme are shown in Figure~\ref{fig:dataset-example-variants}. In the generated dataset, quantities such as variable names, values, collection contents, operand choices, and code lengths are sampled randomly.

\begin{center}\captionsetup{type=figure}
\centering
\begin{tcolorbox}[breakable,
    colback=blue!3,
    colframe=blue!60!black,
    boxrule=0.5pt,
    arc=2pt,
    left=4pt,
    right=4pt,
    top=4pt,
    bottom=4pt,
    width=0.98\textwidth
]
\begin{lstlisting}[basicstyle=\ttfamily\scriptsize,columns=fullflexible,keepspaces=true]
OpenQASM~3;
include "stdgates.inc";
// Complex background 000001
// Contains: timing + classical + openpulse resources
qubit[3] q;
// ---- classical registers / counters (available to core tasks) ----
bit[3] bg_m;
bit[3] bg_m2;
bit[3] syndrome_b;
bool flag;
int[32] i;
int[32] step;
int[32] tries;
int[32] pc;
uint[32] mask;
float[64] gain;
// One generic extern hook for controller-side update logic
extern adaptive_update(int[32], int[32], float[64]) -> float[64];
// ---- timing-friendly constants (core tasks may reuse) ----
const duration BG_IDLE  = 52ns;
const duration RO_RING  = 37us;
const duration BOX_WIN  = 251dt;
defcalgrammar "openpulse";
const float drive_freq_0 = 5.406287e9;
const float meas_freq_0  = 6.109105e9;
const float drive_freq_1 = 4.848509e9;
const float meas_freq_1  = 6.281497e9;
const float drive_freq_2 = 6.744155e9;
const float meas_freq_2  = 6.368663e9;
cal {
  // --- declare ports ---
  extern port d0;
  extern port m0;
  extern port a0;
  extern port d1;
  extern port m1;
  extern port a1;
  extern port d2;
  extern port m2;
  extern port a2;
  // --- extern waveform templates (common) ---
  extern gaussian(complex[float[32]] amp, duration d, duration sigma) -> waveform;
  extern drag(complex[float[32]] amp, duration d, duration sigma, float[32] beta) -> waveform;
  extern constant(complex[float[32]] amp, duration d) -> waveform;
  extern sine(complex[float[32]] amp, duration d, float[64] frequency, angle phase) -> waveform;
  extern gaussian_square(complex[float[32]] amp, duration d, duration square_width, duration sigma) ...
    ... -> waveform;
  // --- extern capture/discriminate hooks (for measure-style tasks) ---
  extern capture(frame capture_frame, waveform filter) -> bit;
  extern capture_v1(frame capture_frame, duration d) -> waveform;
  extern discriminate(complex[float[64]] iq) -> bit;
  // --- create frames ---
  frame q0_drive = newframe(d0, drive_freq_0, 0.0);
  frame q0_meas  = newframe(m0,  meas_freq_0,  0.0);
  frame q0_acq   = newframe(a0,  meas_freq_0,  0.0);
  frame q1_drive = newframe(d1, drive_freq_1, 0.0);
  frame q1_meas  = newframe(m1,  meas_freq_1,  0.0);
  frame q1_acq   = newframe(a1,  meas_freq_1,  0.0);
  frame q2_drive = newframe(d2, drive_freq_2, 0.0);
  frame q2_meas  = newframe(m2,  meas_freq_2,  0.0);
  frame q2_acq   = newframe(a2,  meas_freq_2,  0.0);
  // --- background pulse actions (very lightweight; no play) ---
  shift_phase(q0_drive, 0.266398);
  set_frequency(q0_drive, (get_frequency(q0_drive) + -2611143.0));
  delay[18dt] q0_drive;
  barrier q0_drive, q2_drive;
}
\end{lstlisting}
\end{tcolorbox}
\caption{One example of background program, Part I. This part defines the qubits, classical variables, timing constants, calibration grammar, external pulse-level resources, and frame objects used by downstream core tasks.}
\label{fig:dataset-example-background-part1}
\end{center}

\begin{center}\captionsetup{type=figure}
\centering
\begin{tcolorbox}[breakable,
    colback=blue!3,
    colframe=blue!60!black,
    boxrule=0.5pt,
    arc=2pt,
    left=4pt,
    right=4pt,
    top=4pt,
    bottom=4pt,
    width=0.98\textwidth
]
\begin{lstlisting}[basicstyle=\ttfamily\scriptsize,columns=fullflexible,keepspaces=true]
// --- classical init (lightweight) ---
flag = false;
step = 0;
tries = 0;
pc = 0;
mask = 0;
gain = 1.0;
bg_m[0] = 0;
bg_m2[0] = 0;
syndrome_b[0] = 0;
bg_m[1] = 0;
bg_m2[1] = 0;
syndrome_b[1] = 0;
bg_m[2] = 0;
bg_m2[2] = 0;
syndrome_b[2] = 0;

// --- gate-level + timing background circuit ---
ry(-0.760302) q[0];
rz(1.159948) q[2];
z q[0];
ry(-0.999231) q[1];
rx(-0.237197) q[2];
cz q[1], q[0];
tdg q[0];

// --- timing context: a fixed control window (for DD / readout ring) ---
box[BOX_WIN] {
  delay[BG_IDLE] q[0];
  // independent idle on another qubit to encourage parallel scheduling
  delay[BG_IDLE] q[1];
}

// === CORE_TASK_START ===
// (core complex task will be inserted here)
// === CORE_TASK_END ===

// === MEASUREMENT_START ===
// (measurement block will be inserted here)
// === MEASUREMENT_END ===
\end{lstlisting}
\end{tcolorbox}
\caption{One example of background program, Part II. This part initializes the classical state, provides lightweight gate-level and timing context, and reserves the insertion locations for the core task and the measurement block. In the full dataset, names, values, counts, and related attributes are generated randomly.}
\label{fig:dataset-example-background-part2}
\end{center}

\begin{center}\captionsetup{type=figure}
\centering
\begin{tcolorbox}[breakable,
    colback=blue!3,
    colframe=blue!60!black,
    boxrule=0.5pt,
    arc=2pt,
    left=4pt,
    right=4pt,
    top=4pt,
    bottom=4pt,
    width=0.95\linewidth
]
\begin{lstlisting}[basicstyle=\ttfamily\small,columns=fullflexible,keepspaces=true]
for int __cc_i in {0,1,2,3,4,5} {
  switch (__cc_i % 3) {
    case 0 { x q[2]; }
    case 1 { if (__cc_i > 1) { z q[2]; } else { h q[2]; } }
    default { y q[2]; }
  }
}
\end{lstlisting}
\end{tcolorbox}
\caption{An example core task from the classical-logic category, instantiated from the \texttt{nested\_control\_switch\_in\_loop\_with\_if} theme. As in other dataset instances, the iteration set, operands, and target objects are sampled randomly.}
\label{fig:dataset-example-core-task}
\end{center}

\begin{center}\captionsetup{type=figure}
\centering
\begin{tcolorbox}[breakable,
    colback=blue!3,
    colframe=blue!60!black,
    boxrule=0.5pt,
    arc=2pt,
    left=4pt,
    right=4pt,
    top=4pt,
    bottom=4pt,
    width=0.95\linewidth
]
\begin{lstlisting}[basicstyle=\ttfamily\small,columns=fullflexible,keepspaces=true]
// TODO(core task): A loop runs six iterations with an integer index taking
// values 0 through 5, and in each iteration it applies exactly one
// single-qubit gate to qubit q2 based on the value of the index modulo 3.
// When the index modulo 3 equals 0, it applies an X to q2; 
// When the index modulo 3 equals 1, it applies H to q2 for indices 1 or less, 
// but applies Z to q2 for indices greater than 1. 
// In other conditions, applies Y to q2
\end{lstlisting}
\end{tcolorbox}
\caption{The TODO prompt corresponding to the core task in Figure~\ref{fig:dataset-example-core-task}. This prompt provides a semantic natural-language description of the missing code block.}
\label{fig:dataset-example-todo}
\end{center}

\begin{center}\captionsetup{type=figure}
\centering
\begin{tcolorbox}[breakable,
    colback=blue!3,
    colframe=blue!60!black,
    boxrule=0.5pt,
    arc=2pt,
    left=4pt,
    right=4pt,
    top=4pt,
    bottom=4pt,
    width=0.98\textwidth
]
\begin{lstlisting}[basicstyle=\ttfamily\scriptsize,columns=fullflexible,keepspaces=true]
// Variant 1
for int __cc_i in {0,1,2,3,4,5} {
  switch (__cc_i) {
    case 5 { break; }
    default { x q[2]; }
  }
}

// Variant 2
int __cc_s = 1;
switch (__cc_s) {
  case 0 { x q[0]; }
  default {
    bit __cc_m = measure q[0];
    if (__cc_m) { switch (1) { case 1 { z q[0]; } } } else { h q[0]; }
  }
}

// Variant 3
for int __cc_i in {0,1,2} {
  bit __cc_m = measure q[0];
  int __cc_k = int(__cc_m) + (__cc_i % 2);
  h q[0];
  switch (__cc_k) { case 0 {x q[0];} case 1 {z q[0];} default {h q[0];} }
}
\end{lstlisting}
\end{tcolorbox}
\caption{Three additional variants from the same task theme as Figure~\ref{fig:dataset-example-core-task}. These variants preserve the syntactic and semantic scope of the theme while changing the concrete control-flow structure, operation order, operands, target objects, and parameter choices. As in the full dataset, these elements are sampled randomly.}
\label{fig:dataset-example-variants}
\end{center}
\FloatBarrier

\section{Prompts Used in Experiments}
\label{appendix:prompts}
This section presents all prompts involved in the LLM-based pipeline of this work. They serve three distinct purposes. The first prompt is used to generate variants of core-task generators. The second prompt is used to convert a core task into a natural-language task description. The third prompt is used during evaluation, where the model is asked to generate the missing answer block from the task description and the surrounding program context. For reproducibility, the full prompts are shown in Figures~\ref{fig:prompt-variant-generation}, \ref{fig:prompt-description-generation}, and \ref{fig:prompt-evaluation-completion}.

\begin{center}\captionsetup{type=figure}
\centering
\begin{tcolorbox}[breakable,
    colback=blue!5,
    colframe=blue!60!black,
    boxrule=0.6pt,
    arc=2pt,
    left=6pt,
    right=6pt,
    top=6pt,
    bottom=6pt,
    width=0.97\textwidth,
    fontupper=\small
]
\textbf{Prompt for core-task variant generation}

\medskip
\textbf{Instruction:}

You are an expert in quantum programming.

Below is reference documentation for OpenQASM~3:
\[
\texttt{\{qasm\_documents\}}
\]

Given the following core-task theme
\[
\texttt{\{theme\}}
\]
and the provided core-task template
\[
\texttt{\{template\}},
\]
generate three new core-task generator variants corresponding to versions $v=2,3,4$.

Each variant must remain within the scope of the same theme and use the same syntax and language features as the template, while varying the code structure, statement ordering, and control-flow logic. As in the original template, placeholders should be preserved so that later stages can instantiate them randomly.

For each branch, also provide brief comments and metadata. The comments should indicate which OpenPulse or OpenQASM features are exercised. The metadata should specify any required contextual assumptions, such as the availability of ports or frames, timing granularity in \texttt{dt}, or other hardware-related constraints.
\end{tcolorbox}
\caption{Prompt used to generate new variants of core-task generators from an existing theme and template.}
\label{fig:prompt-variant-generation}
\end{center}

\begin{center}\captionsetup{type=figure}
\centering
\begin{tcolorbox}[breakable,
    colback=blue!5,
    colframe=blue!60!black,
    boxrule=0.6pt,
    arc=2pt,
    left=6pt,
    right=6pt,
    top=6pt,
    bottom=6pt,
    width=0.97\textwidth,
    fontupper=\small
]
\textbf{Prompt for task-description generation}

\medskip
\textbf{System prompt:}

You are given a quantum program snippet. Write a natural-language description of what the snippet does.

Requirements:
\begin{itemize}
    \item Mention the involved objects, such as qubits, classical bits, registers, frames, and durations, together with the key parameters, including angles, indices, durations, and constants.
    \item Do not include code, do not quote any lines, and do not use backticks.
    \item Do not explain how to write the snippet in any programming language, and do not mention punctuation such as brackets, parentheses, or semicolons.
    \item Operations may be described at the conceptual level, for example as rotations, entangling operations, delays, or measurements, but the description should remain semantic and high-level.
    \item Keep the description concise, within 2--6 sentences.
\end{itemize}

\medskip
\textbf{User prompt:}

Context (declarations and earlier operations, for naming only):

\medskip
\texttt{\{background\}}

\medskip
Core snippet to describe (this is the portion to be replaced):

\medskip
\texttt{\{core\_task\}}

\medskip
Following context (may help disambiguate intent):

\medskip
\texttt{\{measurement\_step\}}

\medskip
Now write the description.
\end{tcolorbox}
\caption{Prompt used to convert a core task into a natural-language task description.}
\label{fig:prompt-description-generation}
\end{center}

\begin{center}\captionsetup{type=figure}
\centering
\begin{tcolorbox}[breakable,
    colback=blue!5,
    colframe=blue!60!black,
    boxrule=0.6pt,
    arc=2pt,
    left=6pt,
    right=6pt,
    top=6pt,
    bottom=6pt,
    width=0.97\textwidth,
    fontupper=\small
]
\textbf{Prompt for evaluation-time completion}

\medskip
\textbf{System prompt:}

You complete missing QASM core blocks.

Output only the QASM statements that belong between \texttt{CORE\_TASK\_START} and \texttt{CORE\_TASK\_END}. Do not output the markers themselves. Do not output explanations. Do not use backticks.

\medskip
\textbf{User prompt:}

Here is a QASM program with a missing core block. Fill in the missing core block.

Return only the missing core QASM statements.

\medskip
\texttt{----- BEGIN PROGRAM -----}

\medskip
\texttt{\{prompt\_qasm\}}

\medskip
\texttt{----- END PROGRAM -----}
\end{tcolorbox}
\caption{Prompt used during evaluation to complete the missing core QASM block.}
\label{fig:prompt-evaluation-completion}
\end{center}
\FloatBarrier

\section{Verifier Architecture and Semantic Extensions}
\label{appendix:verifierarchitecture}

The verifier includes an official-parser front end, an OpenQASM~3 interpreter, established quantum-evolution backends, and a resource-aware scheduler. Quantum state/distribution simulations are delegated to existing libraries Qiskit \& Qutip ~\citep{javadiabhari2024qiskit,lambert2026qutip}; the interpreter connects the advanced OpenQASM~3/OpenPulse constructs exercised by the benchmark to those libraries. Reference and candidate programs execute under the same configuration. Acceptance combines task-required constructs with the relevant output-distribution and timeline checks, allowing behavioral comparison without requiring identical source code. In addition, all task themes and their variants were carefully curated so that the verifier's semantic signals (output distributions or timelines) faithfully capture the intended behavior. Semantic success measures satisfaction of these executable task checks, with the background and observation context fixed by the problem.

\paragraph{Unit of evaluation.}
A submission completes one specified block in a supplied program. Its background fixes the objects, parameters, and observation context; the reference provides the corresponding observable behavior. The verifier compares complete joint output records, so dependencies among returned measurement bits enter the score. The scheduling component separately evaluates temporal and resource structure. By fixing the surrounding program and evaluating each completion under the same execution and comparison rules, this design assesses whether models can correctly combine the language features required by each task.

\paragraph{Official-parser front end and AST execution.}
Each complete program is first parsed by the official OpenQASM~3 parser. A candidate must therefore satisfy the language grammar before any benchmark-specific semantic check is applied. The resulting abstract syntax tree (AST) retains declarations, typed expressions, gate invocations, calibration blocks, timing statements, and control-flow structure. We implement an AST visitor that gives operational semantics to the constructs covered by QASM-Eval and maintains a unified runtime state comprising quantum-register mappings, typed classical memory, calibration definitions, ports, frames, waveforms, and timing scopes. The interpreter directly executes arithmetic and bitwise expressions, casts, arrays, assignments, external calls, \texttt{if}/\texttt{else}, \texttt{switch}, \texttt{for}, and \texttt{while}. Mid-circuit measurement collapses the measured subsystem according to the sampled outcome. Subsequent operations follow the corresponding execution branch, enabling measurement-conditioned quantum evolution and classical feed-forward.

\paragraph{Gate-level state semantics through Qiskit.}
The interpreter maps standard one- and multi-qubit operations to Qiskit gate operators and applies them with Qiskit's statevector evolution~\citep{javadiabhari2024qiskit}. Measurement performs state collapse and returns the sampled classical outcome; reset, parameterized gates, calibrated gates, and dynamically selected branches all update the same quantum state. Repeated executions produce an empirical distribution of the complete program's observed outputs, with measurement-conditioned paths sampled during execution. For many quantum computing tasks, the central objective is to produce the intended final quantum state or measurement-outcome distribution. Here, we compare candidate and reference distributions using total variation distance under the same fixed sampling budget and tolerance. Alternative implementations can pass this comparison when their observed distributions agree within the tolerance; the AST element check separately enforces constructs explicitly required by the task.

\paragraph{OpenPulse semantic extension through QuTiP.}
We extend the execution stack with a typed pulse layer that maps OpenPulse objects to an executable control model. Ports identify physical control channels; frames carry their current frequency and phase; and waveform values represent complex envelopes with explicit durations. The layer supports the benchmark's Gaussian, DRAG, constant, sine, Gaussian-square, and sample-array waveforms, as well as waveform mixing, summation, scaling, and phase shifting. Operations such as \texttt{set\_frequency}, \texttt{set\_phase}, and \texttt{shift\_phase} update frame state, and \texttt{play} schedules the resulting envelope on the target frame.

For simulation, the pulse layer converts these objects into a time-dependent lab-frame Hamiltonian
\[
H(t)=H_0+\sum_i\left(c_i(t)\sigma_i^-+c_i^*(t)\sigma_i^+\right),
\]
where $H_0$ contains the configured qubit frequencies and couplings, and $c_i(t)$ combines the waveform envelope, carrier frequency, and accumulated frame phase. QuTiP's time-dependent operator representation and numerical propagator~\citep{lambert2026qutip} compute the corresponding unitary $U(T)$. The unitary is then injected into the same Qiskit statevector used for gate-level operations, allowing pulse blocks, gates, measurements, and classical feedback to compose within one execution. This extension preserves the numerical responsibilities of QuTiP while adding the OpenQASM~3/OpenPulse object semantics and language-to-Hamiltonian mapping required by the benchmark.

\paragraph{Resource-aware timeline semantics.}
State evolution is accompanied by an independent scheduling pass. The scheduler maintains logical clocks for the qubits and frames touched by the program and emits one event for every gate, measurement, reset, delay, pulse playback, barrier, and timing scope. Each event records its start and end times, operation type, resources, and enclosing scope. A configurable system model supplies \texttt{dt}, gate and readout durations, and controller latency, and all quantities are normalized to a common time unit. The scheduler implements duration-valued expressions, \texttt{delay}, \texttt{barrier}, fixed-duration \texttt{box} regions, nested scopes, alignment constraints, and stretch expressions. Multi-resource operations synchronize the corresponding clocks, so the resulting timeline captures both numerical duration and causal/resource ordering. Timing and complex tasks use an event-level comparison that checks operation types, resources, and temporal boundaries against the reference timeline.

\paragraph{Evaluation parameters and robustness.}
The verifier uses 8,000 shots each for the reference and candidate, sampled independently. It should be clarified that these 8,000 shots are the sampling repetitions required when simulating the quantum program, separate from the 5 model generations used to estimate pass@$k$. Distribution-based comparisons use TVD $\leq 0.05$. Pulse propagation uses a 20,001-point time grid to represent the waveform coefficients, with evolution computed by QuTiP. Pulse-task acceptance is determined by the output-distribution criterion, TVD $\leq 0.05$. Independent numerical validation of the pulse backend is reported in Appendix~\ref{appendix:verifier-fid}.

\paragraph{End-to-end decision and validation.}Each submission must parse, contain its task-required constructs, and execute successfully. Classical and pulse tasks then require distribution agreement, timing tasks require timeline agreement, and complex tasks require both, as specified in Section~\ref{subsec: evalmethod}. These checks have distinct roles: the AST check enforces explicitly requested constructs and syntax correctness, execution realizes their effects, and the applicable comparison checks the resulting observations or event trace. For timing tasks, event-level reference agreement is the benchmark's operational criterion, including operation types, resources, and temporal boundaries. Thus, the comparison is explicit and shared across models. Failures are assigned to the detecting layer for the reported error analysis.

\section{Verifier Fidelity}
\label{appendix:verifier-fid}

We assess verifier reliability through reference comparisons and feature-specific validation.

\paragraph{Classical logic tasks.}
We collected 123 common quantum programs (41 tasks, 3 variants for each task) from the official Qiskit and OpenQASM repositories to validate simulation without advanced features. The verifier achieved 100\% agreement with Qiskit, confirming consistency with its underlying simulation backend.
However, classical control flow introduces additional branches whose execution depends on stochastic measurement outcomes. To assess whether the sampling budget adequately supports tasks with these newly-introduced features, we evaluated all classical logic tasks in the test set. The correct-acceptance rate reached 99.9993\% at 8,000 shots, compared with 98.6778\% at 4,000 shots and 83.29\% at 2,000 shots, supporting the selected sampling budget.

\paragraph{Timing tasks.}
Timing evaluation uses a virtual quantum machine with explicitly specified hardware and compilation conventions, including gate durations and default timing intervals. Tasks are bounded in duration and specify the scheduling constraints needed to determine a unique timeline. Our generation pipeline provides detailed timing requirements, and we observed no timeline ambiguity attributable to the natural-language specifications in the examined outputs (Appendix~\ref{appendix:errorexamples}). Across all timing tasks in the test set, the verifier achieved a correct-acceptance rate of 100\%.

\paragraph{Pulse control.}
The pulse backend is implemented using QuTiP. We evaluated 30 pulse configurations spanning waveform families, pulse sequences, phases, and qubit counts, comparing the simulated evolution against an independent numerical reference. All configurations satisfied the prescribed numerical tolerance: the maximum elementwise absolute propagator error was $3.85\times10^{-4}$, and the maximum absolute error in the tested output probabilities was $4.51\times10^{-4}$. These results support accurate pulse evolution across the tested configurations. For binary-output comparisons, the latter error also equals the total variation distance and is approximately 0.90\% of the dataset's decision tolerance of $0.05$.

\paragraph{Disagreements between human experts and the verifier.}
As reported in the main text, the expert audit shows high agreement between human judgments and verifier decisions (94.5\% agreement; Cohen's $\kappa=0.89$). Nevertheless, occasional disagreements reveal differences in how task requirements are interpreted. We examine a representative case below to illustrate how an expert may expect a particular implementation structure, whereas the verifier accepts an alternative that satisfies the explicitly stated requirements and observable behavior.

The following classical-logic task asks the model to apply a sequence of parameterized rotations:

\begin{lstlisting}[basicstyle=\ttfamily\small,columns=fullflexible,keepspaces=true]
// TODO(core task): The snippet defines a base rotation angle of 0.158116 and
// then iterates over the discrete set of integer values 0, 2, 3, 5, and 6.
// For each value, it computes a Z-rotation angle equal to that integer
// multiplied by the base angle, and applies an RZ rotation by that computed
// angle to qubit q[3]. This results in five sequential Z-axis phase rotations
// on q[3], including a zero-angle rotation when the integer is 0.
\end{lstlisting}

One example response generated by LLM was:

\begin{lstlisting}[basicstyle=\ttfamily\small,columns=fullflexible,keepspaces=true]
rz(0.0) q[3];
rz(0.316232) q[3];
rz(0.474348) q[3];
rz(0.79058) q[3];
rz(0.948696) q[3];
\end{lstlisting}

This answer was motivated by the requirement that the rotations be applied sequentially. However, the human expert considered the intended solution to require explicit iteration over the given set, and therefore preferred the following form:

\begin{lstlisting}[basicstyle=\ttfamily\small,columns=fullflexible,keepspaces=true]
angle __cc_base = 0.158116;
for int __cc_i in {0,2,3,5,6} {
  angle __cc_th = angle(float(__cc_i) * float(__cc_base));
  rz(__cc_th) q[3];
}
\end{lstlisting}

The verifier judged the model output to be correct, since the prompt does not explicitly require a loop-based implementation. The human expert, by contrast, applied a stricter interpretation and marked the answer as incorrect. This example shows that disagreement can arise even when the generated code is semantically consistent with the textual description. Accepting the unrolled form follows the prompt's stated rotations while allowing either implementation form.

After reviewing the verifier's decisions and accompanying rationales, the human expert who participated in the agreement study in Section~3.3 concurred with the verifier on the initially disputed cases. The adjudicated false-rejection rate was 0\% (95\% confidence interval: [0\%, 3.36\%]); Table~\ref{tab:verifier-agreement} reports the original independent judgments, which remain the primary agreement measure.

\paragraph{Summary.}
The reference comparisons, feature-specific validation, and high agreement with expert judgments jointly support the verifier's reliability within the benchmark's evaluation scope. Explicit scheduling rules, validated pulse dynamics, and the selected sampling budget provide a consistent basis for assessing the supported task families. The disagreement case further illustrates the distinction between explicit task requirements and implementation preferences, reinforcing the importance of evaluating valid alternative implementations against clearly specified criteria.

\section{Failure Case Study}
\label{appendix:failure-analysis}
\label{appendix:errorexamples}
We present typical syntax and semantic failures from the model-generated programs. The syntax examples show recurring mistakes in writing OpenQASM~3, while the semantic cases illustrate misunderstandings of the quantum processes those programs are meant to implement. We use three examples to show how these misunderstandings change measurement outcomes, qubit states, and interference signals.

\subsection{Common syntax errors}
\label{appendix:syntax-failures}
Syntax errors account for a substantial share of the models' failures. The examples below highlight their common patterns: borrowing grammar from familiar programming languages, confusing instructions with function calls, and using incorrect forms for measurement assignment.
\paragraph{Importing the grammar of a familiar language.}
A recurring pattern is to express controller operations with punctuation from C-like languages. Examples include a colon after a case label and a three-part loop header. The following excerpts normalize variable names and values to isolate the grammatical distinction:
\begin{lstlisting}[basicstyle=\ttfamily\small,columns=fullflexible,keepspaces=true]
// Invalid C-style case label       // OpenQASM 3 form
switch (s) {                       switch (s) {
  case 0: { x q[0]; }                case 0 { x q[0]; }
}                                  }

// Invalid C-style loop header
for (int i = 0; i < 3; i++) { x q[0]; }
// OpenQASM 3 iteration over the specified values
for int i in {0, 1, 2} { x q[0]; }
\end{lstlisting}
Both substitutions retain a recognizable branch or loop structure. Execution requires the exact grammar of the target language, including its case bodies and iteration syntax~\citep{cross2022openqasm3}.

\paragraph{Confusing an instruction with a function call.}
Some responses express a delay as a conventional function call. In OpenQASM~3, its duration belongs in brackets and its resource operands follow the instruction:
\begin{lstlisting}[basicstyle=\ttfamily\small,columns=fullflexible,keepspaces=true]
// Incorrect form for the requested timing instruction
 delay(40ns, q[0]);
// OpenQASM 3 timing instruction
 delay[40ns] q[0];
\end{lstlisting}
The duration-resource pair specifies which resource clock advances. The generated function-call form transfers familiar syntax to a specialized timing instruction.

\paragraph{Measurement assignment and identifier boundaries.}
Other responses use \texttt{to} or \texttt{>} as measurement connectors, or reuse a reserved type token as a variable name. A valid measurement assignment is \texttt{m = measure q[0];}; the receiving identifier must also satisfy the language's naming rules. These examples locate failures at the interface between measurement instructions and ordinary classical variables.

\subsection{Semantic failures}
\label{appendix:semantic-failures}
The next examples show that LLMs still lack understanding of quantum operations and processes, and also show how a plausible-looking completion can get the underlying quantum process wrong. They concern measurement, pulse control, and spin echo. In each case, we explain the intended behavior, identify the model's misunderstanding, and use the figures to show how it changes the result.

\subsubsection{Treating a basis rotation as a prerequisite for measurement}
\paragraph{Case explanation.}
The first task asks the model to measure a qubit three times and apply a conditional rotation after each measurement. Each round uses a different angle. In the reference program, outcome zero selects an $R_x$ rotation and outcome one selects an $R_z$ rotation. The angles determine the feedback applied after the result is obtained. Measurement itself takes place in the computational basis, as specified by the OpenQASM measurement instruction~\citep{cross2022openqasm3}.

This order matters because measurement supplies the information used by the next operation. For an incoming state $\rho$, the probability of result $m$ is $P(m)=\operatorname{Tr}(\Pi_m\rho)$, where $\Pi_m=|m\rangle\langle m|$. The corresponding conditional rotation then acts on the post-measurement state. In this example, once a measurement returns one, the selected $R_z$ preserves $|1\rangle$ up to phase, so the next measurement also returns one. The program therefore carries information from one round into the next through both the measured bit and the qubit's state.

A rotation before measurement serves a different purpose: it changes the basis being measured. For example, adding $R_y(\theta)$ makes the measurement equivalent, in its outcome probabilities, to measuring the observable
\begin{equation}
M_\theta=R_y^\dagger(\theta)ZR_y(\theta)
=\cos\theta\,Z-\sin\theta\,X
\label{eq:fj-measurement-axis}
\end{equation}
on the incoming state. Such a rotation is useful when a different measurement basis is requested. Here, the task calls for measuring the existing state and using the result to choose the subsequent rotation.

\begin{figure}[htbp]
\centering
\includegraphics[width=\linewidth]{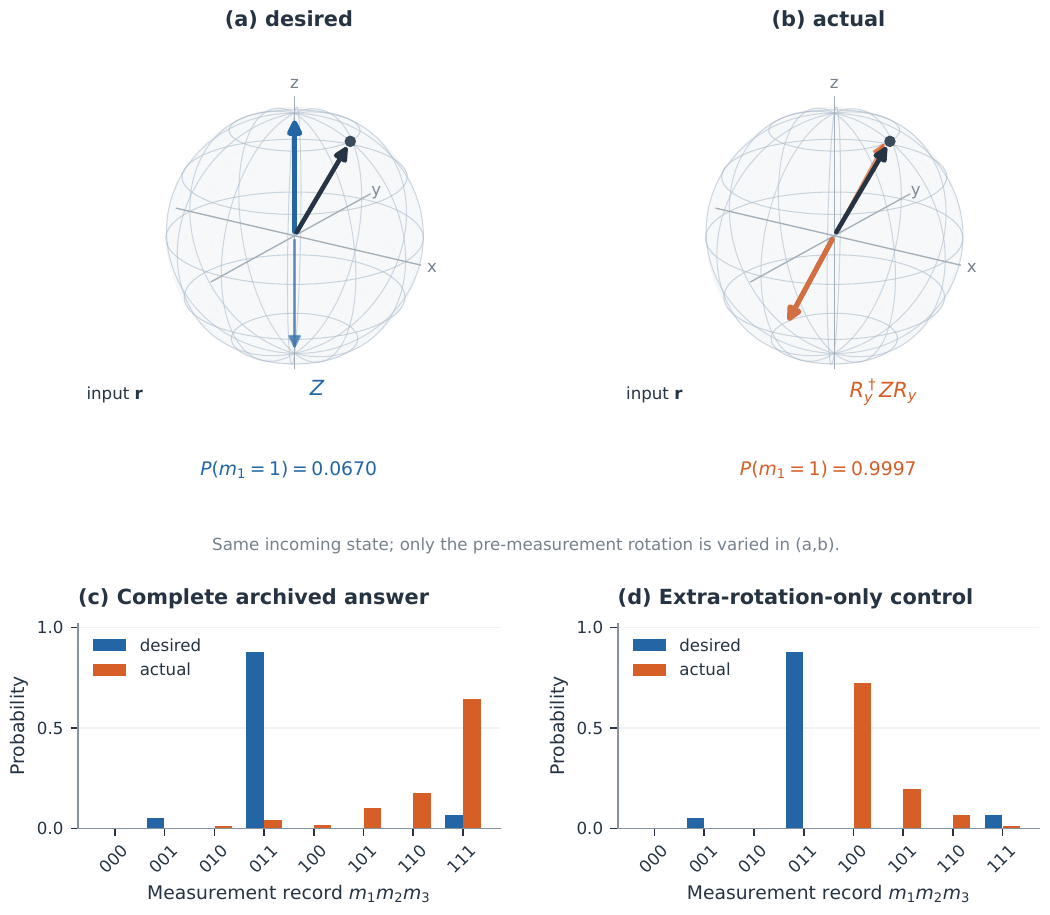}
\caption{An unnecessary rotation changes what is measured. (a,b) The same incoming state (dark vector) is measured along the desired or altered axis (colored vectors). The positive axis corresponds to outcome zero and the opposite direction to outcome one. These panels vary only the extra $R_y$ rotation. (c) Three-round measurement records for the complete desired and actual programs. (d) The same comparison with only the extra rotations added to the reference, keeping its initial state and feedback branches. Probabilities are calculated from the full five-qubit state of C18/v4/instance2.}
\label{fig:fj-measurement-process}
\end{figure}

\paragraph{LLM's failure.}
The model treats rotation into a particular basis as a prerequisite for a valid measurement. It therefore adds $R_y(\theta_j)$ before every measurement, borrowing the angle intended for the later feedback operation. This mistake appears in one of DPSK-V4-Pro's answers to task C18. The model also resets the qubit at entry and exchanges the reference feedback branches; Fig.~\ref{fig:fj-measurement-process} separates the effect of the extra rotations from the complete answer.

The Bloch spheres in panels (a,b) show what the extra rotation does to the first measurement. Both panels use the same incoming state, whose Bloch vector is approximately $(0.5,0,0.8660)$. Measuring it directly gives outcome one with probability $0.0670$. With the inserted rotation, the effective measurement axis points almost opposite to the state, and that probability rises to $0.9997$. The model's supposed preparation for measurement has changed the information that the measurement extracts.

That change continues through the feedback loop. Each result selects the next rotation, and each rotated state becomes the input to the next measurement. Panel (d) shows the consequence of adding only the pre-measurement rotations: the most likely three-round record changes from $011$ to $100$. Panel (c) shows the complete generated answer, for which $111$ becomes the most likely record. After the third feedback operation, the final probability of measuring one is $0.99929$ for the desired program and $0.12720$ for the complete answer. These results come from exact evolution of the supplied five-qubit program, including all measurement branches. The case illustrates a weakness in distinguishing measurement-basis selection from measurement-driven feedback: an unnecessary operation before measurement changes the decisions made throughout the program.

\subsubsection{Mistaking a waveform's name for its rotation axis}
\paragraph{Case explanation.}
The second task asks for an XY4-like sequence: four pulses that alternate between the $X$ and $Y$ axes. Such control sequences are used in dynamical decoupling to protect quantum coherence~\citep{khodjasteh2005DD}. In the ideal $\pi$-pulse limit, the four rotations complete a cycle and return the qubit to its starting state up to a global phase:
\begin{equation}
R_y(\pi)R_x(\pi)R_y(\pi)R_x(\pi)=-I.
\end{equation}

The task defines two waveforms with identical real Gaussian envelopes. Their different rotation axes come from the phase of the drive frame when each waveform is played. For an envelope $\Omega(t)$ and phase $\phi$, the two drive quadratures and the corresponding Hamiltonian are
\begin{equation}
I(t)=\Omega(t)\cos\phi,\quad Q(t)=\Omega(t)\sin\phi,
\qquad H_d(t)=\tfrac12[I(t)X+Q(t)Y],
\label{eq:fj-pulse-axis}
\end{equation}
with $\hbar=1$. A phase of zero gives an $X$ rotation; a phase of $\pi/2$ gives a $Y$ rotation. OpenPulse retains the frame phase between waveform plays~\citep{cross2022openqasm3}. The program must therefore update that phase whenever it switches between the two axes, including before the final $Y$ pulse.

\paragraph{LLM's failure.}
The model treats the waveform named for a $Y$ pulse as if the name itself made the pulse rotate about $Y$. In one of Llama-70B' answers to task X20, it correctly changes the phase before pulses two and three, but omits the update before pulse four. At that point the frame has returned to phase zero. Playing the waveform labeled as $Y$ therefore produces another $X$ rotation, giving XYXX.

\begin{figure}[!htbp]
\centering
\includegraphics[width=\linewidth]{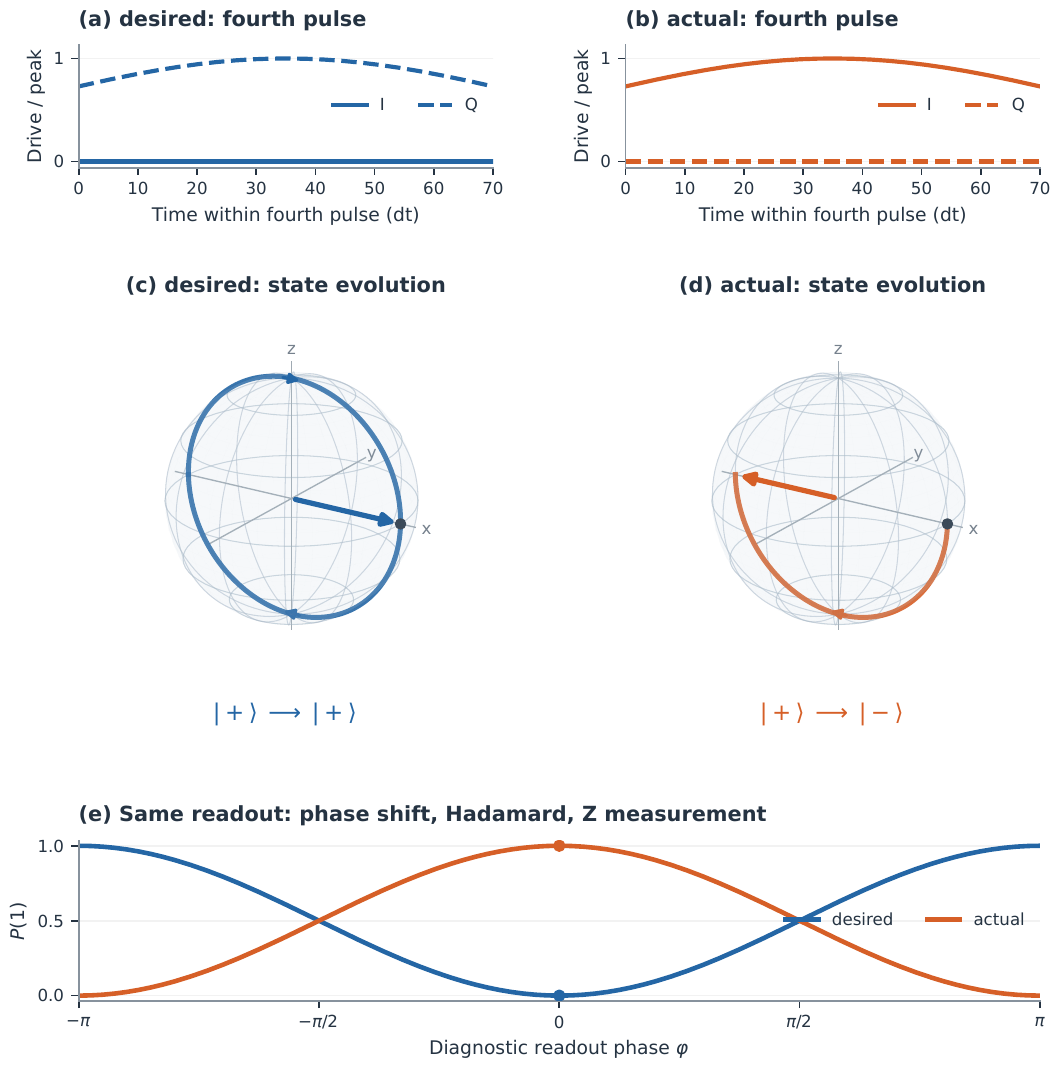}
\caption{A missing phase update changes the rotation axis. (a,b) I/Q components of the fourth Gaussian pulse, relative to the incoming drive frame and normalized to the peak. Solid and dashed lines denote I and Q. (c,d) Bloch trajectories starting from $|+\rangle$; the dark point marks the input and the radial arrow the output. (e) A common phase-sensitive readout gives opposite interference fringes. This illustration uses the phase-update pattern from X20, idealized multiples of $\pi/2$, and calibrated ideal $\pi$ pulses to isolate the axis error from other scheduling changes. The Gaussian envelope has duration 70\,dt and width 44\,dt.}
\label{fig:fj-xy4-process}
\end{figure}

Figure~\ref{fig:fj-xy4-process}(a,b) shows the physical difference in the final pulse. Its Gaussian envelope has the same shape in both answers, but the desired drive is in the $Q$ quadrature and the actual drive is in $I$. This changes the rotation axis and, consequently, the state at the end of the sequence. With calibrated ideal $\pi$ pulses, the actual sequence gives
\begin{equation}
R_x(\pi)R_x(\pi)R_y(\pi)R_x(\pi)=-iZ.
\end{equation}
Instead of returning the input state, it introduces a relative sign between the $|0\rangle$ and $|1\rangle$ components. The Bloch trajectories in panels (c,d) follow a common $|+\rangle$ input: the desired path closes at $|+\rangle$, while the actual path stops at the opposite state, $|-\rangle$.

This phase error affects the result of a later computation. Panel (e) applies the same readout to both states: a phase rotation $R_z(\varphi)$, followed by a Hadamard and computational-basis measurement. The outcome probabilities are
\begin{equation}
P_{\mathrm{desired}}(1)=\sin^2(\varphi/2),\qquad
P_{\mathrm{actual}}(1)=\cos^2(\varphi/2).
\end{equation}
At zero readout phase, the desired program returns zero and the actual program returns one. The opposite interference fringes show how an omitted frame update turns a state-preserving cycle into a coherent error. This case highlights the model's difficulty connecting a pulse description to its physical action through the current drive phase.

\subsubsection{Expecting a refocusing pulse to complete an echo by itself}
\paragraph{Case explanation.}
The third task asks for a Hahn echo. After a preparation pulse, the qubit evolves freely for $\tau=140\,\mathrm{dt}$, receives a $\pi$ refocusing pulse, and evolves for another $\tau$ before the final analysis pulse. The two waiting periods allow the sequence to cancel phase accumulated from a slowly varying detuning~\citep{bylander2011DD_timing}.

During the first interval, different detunings cause spins to accumulate different phases, spreading their Bloch vectors around the equator. The refocusing pulse reverses their relative phase evolution. Over the second interval, they come back together, recovering a well-defined phase when the analysis pulse arrives. For quasi-static detuning $\delta$, this follows from
\begin{equation}
U_z(t_2)XU_z(t_1)=XU_z(t_1-t_2),
\qquad U_z(t)=e^{-i\delta tZ/2}.
\label{eq:fj-echo}
\end{equation}
With equal intervals, $t_1=t_2$, the detuning-dependent phase cancels. The final analysis pulse converts the recovered coherence into a population that can be measured. Its position therefore matters: it should arrive after the spins have had time to refocus.

\paragraph{LLM's failure.}
The model treats the $\pi$ pulse as if it completed the refocusing on its own. In Llama-8B's fifth answer to X03, it keeps the three pulses and the first waiting period but places the analysis pulse immediately after the refocusing pulse. The second 140\,dt interval is missing. The answer still includes a later padding delay, but by then the analysis pulse has already converted phase into population. Waiting at that point cannot recover the signal lost by analyzing the spins too early.

\begin{figure}[!htbp]
\centering
\includegraphics[width=\linewidth]{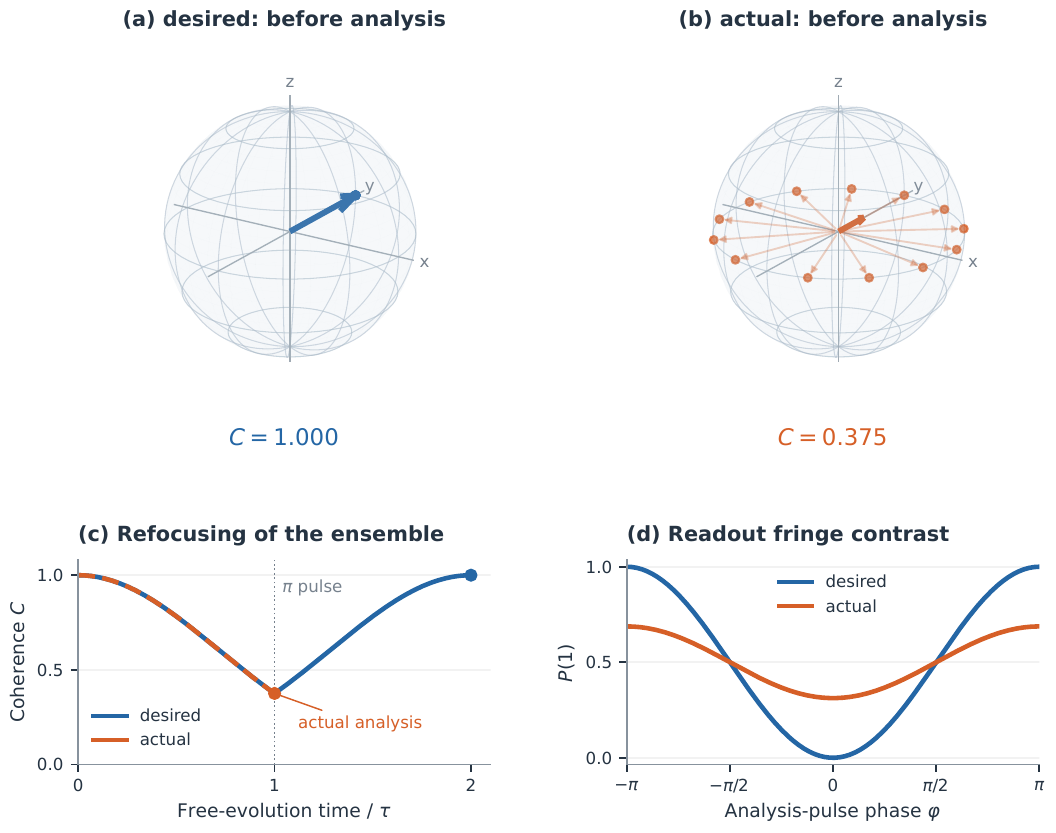}
\caption{Reading out before the spins have refocused. The illustration isolates the missing second wait in X03, using ideal pulses and Gaussian quasi-static detuning with $\sigma_\delta\tau=1.4$. (a,b) Bloch vectors just before analysis. Thin vectors represent individual detunings; the thick vector is the Gaussian-weighted average. (c) The desired coherence recovers during the second interval; the actual sequence applies its analysis pulse before this recovery. (d) The resulting reduction in readout-fringe contrast. The same $R_x(\pi/2)$ preparation, $R_x(\pi)$ refocusing, and analysis-phase scan are used in both sequences.}
\label{fig:fj-echo-process}
\end{figure}

Figure~\ref{fig:fj-echo-process} shows the resulting loss of coherence. Both sequences start in $|0\rangle$ and use the same ideal preparation and refocusing pulses. Across repetitions, the detuning follows a zero-mean Gaussian distribution with $\sigma_\delta\tau=1.4$. Immediately before analysis, the desired Bloch vectors have come back together at $+y$, while the actual vectors remain spread around the equator. The length of their average transverse vector is
\begin{equation}
C=\exp\!\left[-\tfrac12\sigma_\delta^2(t_1-t_2)^2\right].
\end{equation}
The desired sequence restores $C=1$; omitting the second wait leaves $C\approx0.375$. Panel (c) shows precisely where the actual sequence stops the recovery by applying its analysis pulse.

The readout makes this difference observable. Scanning the phase $\varphi$ of the final $\pi/2$ pulse gives $P(1\mid\varphi)=[1-C\cos\varphi]/2$. The desired signal spans zero to one, while the actual signal spans only about $0.312$ to $0.688$, as shown in panel (d). The reduced contrast means that much of the phase information the echo should recover is lost from the measured signal. This case shows the model's difficulty understanding free evolution as an essential part of a quantum protocol: the right pulses alone do not produce an echo when the interval needed for refocusing is missing.

\FloatBarrier

\section{Detailed Model Performance}
\label{appendix:detailed-performance}
We report pass@1 through pass@5 for all configurations, first over all 1,200 tasks and then within each 300-task category. Model names and suffixes follow the main text. All rates are percentages. Each curve uses the same five archived generations per task. For a task with $c_i$ accepted generations among $n=5$, the estimator is
\begin{equation}
\widehat{\mathrm{pass@}k}=\frac{1}{N}\sum_{i=1}^{N}\left[1-\frac{\binom{5-c_i}{k}}{\binom{5}{k}}\right],\qquad k=1,\ldots,5,
\end{equation}
with $\binom{a}{k}=0$ when $a<k$. Pass@1 is the mean acceptance rate, and pass@5 is the fraction of tasks with at least one accepted generation in the five-output set. Syntax is measured over individual generations. The marginal gain from one budget to the next measures the additional task coverage available when more candidates can be checked.

\subsection{Overall profile: first-attempt reliability and sampling gains}
\begin{table}[htbp]
\centering\small
\caption{Overall pass@$k$ and syntax-valid rates (\%). $\Delta$ is pass@5 minus pass@1 in percentage points.}
\label{tab:v5-overall}
{\fontsize{9}{11}\selectfont\setlength{\tabcolsep}{3pt}\renewcommand{\arraystretch}{1.13}
\begin{tabular*}{\linewidth}{@{\extracolsep{\fill}}lrrrrrrr@{}}
\toprule
Model & pass@1 & pass@2 & pass@3 & pass@4 & pass@5 & Syntax & $\Delta$ \\ \midrule
GPT-5.6-Terra & 39.65 & 44.73 & 47.09 & 48.50 & 49.50 & 67.03 & 9.85 \\
DPSK-V4-Pro & 25.67 & 31.16 & 34.28 & 36.55 & 38.25 & 54.03 & 12.58 \\
Llama-70B-base & 15.70 & 17.39 & 18.13 & 18.57 & 18.83 & 32.77 & 3.13 \\
Llama-8B-base & 9.98 & 13.65 & 15.91 & 17.50 & 18.67 & 40.50 & 8.68 \\
GPT-5.6-Terra-Agent & 47.37 & 51.59 & 53.36 & 54.45 & 55.25 & 75.72 & 7.88 \\
GPT-5.6-Terra-fs & 63.07 & 68.31 & 69.82 & 70.63 & 71.17 & 93.57 & 8.10 \\
DPSK-V4-Pro-fs & 60.10 & 66.95 & 69.24 & 70.62 & 71.58 & 93.47 & 11.48 \\
Llama-70B-fs & 55.28 & 62.84 & 65.94 & 67.92 & 69.42 & 81.82 & 14.13 \\
Llama-8B-fs & 32.87 & 42.73 & 47.74 & 50.93 & 53.25 & 63.85 & 20.38 \\
Llama-70B-QCB & 17.33 & 29.27 & 37.88 & 44.38 & 49.50 & 39.53 & 32.17 \\
Llama-8B-QCB & 8.67 & 15.75 & 21.62 & 26.52 & 30.67 & 28.52 & 22.00 \\
Llama-70B-AGQ & 16.05 & 27.44 & 35.74 & 41.97 & 46.75 & 35.53 & 30.70 \\
Llama-8B-AGQ & 5.83 & 11.01 & 15.62 & 19.73 & 23.42 & 23.20 & 17.58 \\
Llama-70B-ours & 61.17 & 67.00 & 69.69 & 71.50 & 72.92 & 92.57 & 11.75 \\
Llama-8B-ours & 46.93 & 53.79 & 56.92 & 58.85 & 60.25 & 90.38 & 13.32 \\
\bottomrule
\end{tabular*}}
\end{table}
Table~\ref{tab:v5-overall} separates the quality of a typical generation from the chance of obtaining an accepted completion within a five-sample budget. GPT-5.6-Terra-fs has the highest pass@1, at 63.07\%, while Llama-70B-ours reaches the highest pass@5, at 72.92\%. The crossover appears between pass@3 and pass@4: the two configurations score 69.82\% and 69.69\% at three samples, then 70.63\% and 71.50\% at four. This distinction matters for the generation workflow. A single-completion setting rewards reliable first attempts; a verifier-guided workflow can exploit additional candidates.

The shape of the curve reveals where those additional candidates help. GPT-5.6-Terra-fs gains 5.24 percentage points from the second sample and 0.54 from the fifth. Llama-70B-ours gains 5.83 and 1.42 points at the corresponding steps. Its later samples continue to add coverage more rapidly, even though its first-attempt rate is slightly lower. The comparison identifies two different strengths: concentrated success at small budgets and broader recovery as the candidate budget increases.

The QCB and AGQ configurations have a different profile. Llama-70B-QCB rises from 17.33\% to 49.50\%, and Llama-70B-AGQ from 16.05\% to 46.75\%. Their large sampling gains coexist with low first-attempt accuracy. For the estimator above, a task with one accepted output among five contributes 0.2 at pass@1 and 1 at pass@5, whereas a task with five accepted outputs contributes 1 throughout. Large gains therefore locate success in tasks with mixed accepted and rejected generations. QASM-Eval adaptation gives Llama-70B-ours both a stronger initial rate and higher final coverage, reducing the candidate budget needed to reach a given acceptance level.

The overall average combines categories with markedly different profiles. To understand which quantum-programming requirements remain difficult, the following tables retain the complete sampling curve within each category.

\subsection{Classical control: preserving the measurement-conditioned process}
\begin{table}[htbp]
\centering\small
\caption{Classical category pass@$k$ and syntax-valid rates (\%). Each configuration is evaluated on 300 tasks with five generations per task.}
\label{tab:v5-classical}
{\fontsize{9}{11}\selectfont\setlength{\tabcolsep}{3pt}\renewcommand{\arraystretch}{1.13}
\begin{tabular*}{\linewidth}{@{\extracolsep{\fill}}lrrrrrr@{}}
\toprule
Model & pass@1 & pass@2 & pass@3 & pass@4 & pass@5 & Syntax \\ \midrule
GPT-5.6-Terra & 30.87 & 37.67 & 40.67 & 42.33 & 43.33 & 70.80 \\
DPSK-V4-Pro & 22.67 & 30.03 & 33.93 & 36.67 & 38.67 & 82.07 \\
Llama-70B-base & 13.53 & 16.73 & 18.03 & 18.80 & 19.33 & 39.20 \\
Llama-8B-base & 6.33 & 9.60 & 11.77 & 13.27 & 14.33 & 33.47 \\
GPT-5.6-Terra-Agent & 42.00 & 49.27 & 52.27 & 54.07 & 55.33 & 91.93 \\
GPT-5.6-Terra-fs & 47.87 & 55.80 & 58.73 & 60.20 & 61.00 & 88.27 \\
DPSK-V4-Pro-fs & 45.00 & 54.67 & 58.43 & 60.60 & 62.00 & 90.60 \\
Llama-70B-fs & 31.20 & 38.70 & 42.63 & 45.27 & 47.33 & 63.80 \\
Llama-8B-fs & 34.40 & 43.20 & 47.37 & 50.13 & 52.33 & 82.67 \\
Llama-70B-QCB & 12.33 & 22.83 & 31.77 & 39.33 & 45.67 & 39.40 \\
Llama-8B-QCB & 2.93 & 5.73 & 8.40 & 10.93 & 13.33 & 19.27 \\
Llama-70B-AGQ & 10.33 & 19.20 & 26.77 & 33.20 & 38.67 & 34.53 \\
Llama-8B-AGQ & 2.07 & 4.10 & 6.10 & 8.07 & 10.00 & 15.73 \\
Llama-70B-ours & 36.07 & 42.73 & 46.00 & 48.20 & 50.00 & 80.53 \\
Llama-8B-ours & 37.07 & 48.07 & 53.47 & 56.60 & 58.67 & 86.13 \\
\bottomrule
\end{tabular*}}
\end{table}
Table~\ref{tab:v5-classical} shows substantial sampling gains even for the stronger configurations. GPT-5.6-Terra-fs improves from 47.87\% to 61.00\%, and DPSK-V4-Pro-fs from 45.00\% to 62.00\%. Llama-8B-ours rises from 37.07\% to 58.67\%, with gains of 11.00 points from the second candidate and a further 10.60 points across the remaining three. The first-attempt ranking consequently understates the five-candidate coverage of the adapted smaller model.

These tasks combine controller logic with operations on a state that changes after measurement. The syntax-valid rates of GPT-5.6-Terra-fs, DPSK-V4-Pro-fs, and Llama-8B-ours are all above 86\%, while their pass@1 rates remain below 48\%. The separation directs attention to the behavior expressed by valid programs: which measured bit selects an operation, how the conditional state enters the next loop iteration, and when the process stops. The state-reset case in Appendix~\ref{appendix:semantic-failures} makes this requirement concrete. A plausible preparation routine can change the trajectory requested by the task.

Llama-8B-ours also exceeds Llama-70B-ours at every reported sampling budget in this category, with pass@5 of 58.67\% versus 50.00\%. The category thus isolates a relative strength of the smaller adapted model that is obscured by the larger model's stronger overall score. For further improvement, these results motivate examples and feedback that follow the joint classical record and conditional quantum state through a complete control path.

\subsection{Timing: deriving the schedule and positioning its events}
\begin{table}[htbp]
\centering\small
\caption{Timing category pass@$k$ and syntax-valid rates (\%). Each configuration is evaluated on 300 tasks with five generations per task.}
\label{tab:v5-timing}
{\fontsize{9}{11}\selectfont\setlength{\tabcolsep}{3pt}\renewcommand{\arraystretch}{1.13}
\begin{tabular*}{\linewidth}{@{\extracolsep{\fill}}lrrrrrr@{}}
\toprule
Model & pass@1 & pass@2 & pass@3 & pass@4 & pass@5 & Syntax \\ \midrule
GPT-5.6-Terra & 64.00 & 68.63 & 70.60 & 71.67 & 72.33 & 84.33 \\
DPSK-V4-Pro & 39.00 & 44.37 & 47.83 & 50.40 & 52.33 & 69.53 \\
Llama-70B-base & 24.40 & 25.17 & 25.53 & 25.80 & 26.00 & 54.87 \\
Llama-8B-base & 12.40 & 15.10 & 16.53 & 17.53 & 18.33 & 71.13 \\
GPT-5.6-Terra-Agent & 78.67 & 81.20 & 82.23 & 82.87 & 83.33 & 95.27 \\
GPT-5.6-Terra-fs & 85.13 & 87.83 & 88.20 & 88.47 & 88.67 & 97.20 \\
DPSK-V4-Pro-fs & 81.27 & 86.50 & 87.80 & 88.67 & 89.33 & 96.93 \\
Llama-70B-fs & 75.00 & 83.80 & 87.43 & 89.73 & 91.33 & 91.47 \\
Llama-8B-fs & 52.27 & 64.53 & 69.90 & 73.00 & 75.00 & 87.07 \\
Llama-70B-QCB & 17.20 & 31.17 & 42.43 & 51.53 & 59.00 & 38.67 \\
Llama-8B-QCB & 11.47 & 21.37 & 29.93 & 37.40 & 44.00 & 55.87 \\
Llama-70B-AGQ & 18.00 & 32.43 & 43.93 & 53.07 & 60.33 & 36.80 \\
Llama-8B-AGQ & 8.40 & 16.00 & 22.90 & 29.20 & 35.00 & 49.60 \\
Llama-70B-ours & 85.33 & 91.10 & 93.30 & 94.53 & 95.33 & 97.20 \\
Llama-8B-ours & 54.20 & 60.30 & 63.10 & 65.07 & 66.67 & 92.60 \\
\bottomrule
\end{tabular*}}
\end{table}
The timing profiles in Table~\ref{tab:v5-timing} distinguish repeatable scheduling from schedules recovered through additional attempts. Llama-70B-ours and GPT-5.6-Terra-fs begin almost level at 85.33\% and 85.13\%, but reach 95.33\% and 88.67\% at pass@5. Their syntax-valid rates are both 97.20\%. The larger final separation therefore appears after the language form has largely been mastered, within the task's duration and resource constraints.

The comparison between Llama-8B-fs and Llama-8B-ours further illustrates why all five budgets are useful. QASM-Eval adaptation raises syntax validity from 87.07\% to 92.60\% and pass@1 from 52.27\% to 54.20\%. The few-shot configuration has greater five-sample coverage, reaching 75.00\% versus 66.67\%. Its second candidate adds 12.26 points, compared with 6.10 for the adapted configuration. The practical distinction is between improving the average generated schedule and increasing the variety of tasks for which sampling yields an accepted schedule.

The physical requirement is to locate each control event on the appropriate resource timeline. Duration arithmetic, shared stretch variables, and nested windows must agree across the whole fragment. This explains the value of checking the schedule directly: the Hahn-echo case in Appendix~\ref{appendix:semantic-failures} preserves recognizable pulses but loses the waiting interval needed before the final analysis pulse. For timing tasks, useful feedback identifies the misplaced event and the constraint it should satisfy, so the model can revise the relation between operations as well as their numerical durations.

\subsection{Pulse control: strong local generation and the role of sampling}
\begin{table}[htbp]
\centering\small
\caption{Pulse category pass@$k$ and syntax-valid rates (\%). Each configuration is evaluated on 300 tasks with five generations per task.}
\label{tab:v5-pulse}
{\fontsize{9}{11}\selectfont\setlength{\tabcolsep}{3pt}\renewcommand{\arraystretch}{1.13}
\begin{tabular*}{\linewidth}{@{\extracolsep{\fill}}lrrrrrr@{}}
\toprule
Model & pass@1 & pass@2 & pass@3 & pass@4 & pass@5 & Syntax \\ \midrule
GPT-5.6-Terra & 61.40 & 68.83 & 72.37 & 74.60 & 76.33 & 66.40 \\
DPSK-V4-Pro & 40.07 & 48.83 & 53.77 & 57.47 & 60.33 & 43.33 \\
Llama-70B-base & 24.53 & 27.33 & 28.63 & 29.33 & 29.67 & 28.73 \\
Llama-8B-base & 21.20 & 29.90 & 35.33 & 39.20 & 42.00 & 42.80 \\
GPT-5.6-Terra-Agent & 64.67 & 70.23 & 72.40 & 73.73 & 74.67 & 69.47 \\
GPT-5.6-Terra-fs & 90.60 & 95.80 & 96.70 & 97.07 & 97.33 & 97.07 \\
DPSK-V4-Pro-fs & 88.07 & 93.90 & 94.63 & 94.87 & 95.00 & 95.33 \\
Llama-70B-fs & 81.87 & 89.60 & 91.37 & 92.27 & 93.00 & 85.40 \\
Llama-8B-fs & 41.27 & 56.97 & 65.50 & 70.87 & 74.67 & 47.93 \\
Llama-70B-QCB & 39.00 & 61.47 & 74.90 & 83.47 & 89.33 & 65.27 \\
Llama-8B-QCB & 20.07 & 35.50 & 47.53 & 56.93 & 64.33 & 32.73 \\
Llama-70B-AGQ & 35.27 & 56.93 & 70.47 & 79.20 & 85.00 & 59.20 \\
Llama-8B-AGQ & 12.73 & 23.67 & 33.07 & 41.13 & 48.00 & 23.33 \\
Llama-70B-ours & 89.33 & 92.43 & 93.67 & 94.47 & 95.00 & 96.27 \\
Llama-8B-ours & 83.80 & 89.33 & 90.90 & 91.73 & 92.33 & 91.87 \\
\bottomrule
\end{tabular*}}
\end{table}
Table~\ref{tab:v5-pulse} shows the strongest category-level performance for many configurations. GPT-5.6-Terra-fs reaches 90.60\% at pass@1 and 97.33\% at pass@5; Llama-70B-ours reaches 89.33\% and 95.00\%. Llama-8B-ours reaches 83.80\% on the first attempt, improving sharply over the 41.27\% of Llama-8B-fs. At five candidates the corresponding rates are 92.33\% and 74.67\%. The adapted model therefore improves both the typical completion and the set of tasks covered by the available samples.

The high-performing curves flatten quickly. GPT-5.6-Terra-fs gains 5.20 points from the second candidate, followed by 0.90, 0.37, and 0.26 points. Once a configuration reliably produces the requested pulse operation, more candidates offer diminishing additions to coverage. For such a setting, inspecting the remaining failures and supplying the missing calibration or frame relationship becomes a more targeted next step than uniformly increasing the sampling budget.

Llama-70B-QCB exhibits a complementary pattern: its pass@1 is 39.00\%, yet pass@5 reaches 89.33\%. Llama-70B-AGQ similarly moves from 35.27\% to 85.00\%. These configurations frequently find an accepted pulse completion within the candidate set while producing a lower average acceptance rate. Their pulse results account for a substantial part of their overall sampling gains. The complex-category results below locate the next challenge: using waveform construction and frame manipulation inside longer feedback and timing workflows.

\subsection{Complex workflows: composing state, frame, and timing dependencies}
\begin{table}[htbp]
\centering\small
\caption{Complex category pass@$k$ and syntax-valid rates (\%). Each configuration is evaluated on 300 tasks with five generations per task.}
\label{tab:v5-complex}
{\fontsize{9}{11}\selectfont\setlength{\tabcolsep}{3pt}\renewcommand{\arraystretch}{1.13}
\begin{tabular*}{\linewidth}{@{\extracolsep{\fill}}lrrrrrr@{}}
\toprule
Model & pass@1 & pass@2 & pass@3 & pass@4 & pass@5 & Syntax \\ \midrule
GPT-5.6-Terra & 2.33 & 3.80 & 4.73 & 5.40 & 6.00 & 46.60 \\
DPSK-V4-Pro & 0.93 & 1.40 & 1.60 & 1.67 & 1.67 & 21.20 \\
Llama-70B-base & 0.33 & 0.33 & 0.33 & 0.33 & 0.33 & 8.27 \\
Llama-8B-base & 0.00 & 0.00 & 0.00 & 0.00 & 0.00 & 14.60 \\
GPT-5.6-Terra-Agent & 4.13 & 5.67 & 6.53 & 7.13 & 7.67 & 46.20 \\
GPT-5.6-Terra-fs & 28.67 & 33.80 & 35.63 & 36.80 & 37.67 & 91.73 \\
DPSK-V4-Pro-fs & 26.07 & 32.73 & 36.10 & 38.33 & 40.00 & 91.00 \\
Llama-70B-fs & 33.07 & 39.27 & 42.33 & 44.40 & 46.00 & 86.60 \\
Llama-8B-fs & 3.53 & 6.20 & 8.20 & 9.73 & 11.00 & 37.73 \\
Llama-70B-QCB & 0.80 & 1.60 & 2.40 & 3.20 & 4.00 & 14.80 \\
Llama-8B-QCB & 0.20 & 0.40 & 0.60 & 0.80 & 1.00 & 6.20 \\
Llama-70B-AGQ & 0.60 & 1.20 & 1.80 & 2.40 & 3.00 & 11.60 \\
Llama-8B-AGQ & 0.13 & 0.27 & 0.40 & 0.53 & 0.67 & 4.13 \\
Llama-70B-ours & 33.93 & 41.73 & 45.80 & 48.80 & 51.33 & 96.27 \\
Llama-8B-ours & 12.67 & 17.47 & 20.20 & 22.00 & 23.33 & 90.93 \\
\bottomrule
\end{tabular*}}
\end{table}
The complex category in Table~\ref{tab:v5-complex} remains the clearest concentration of difficulty. The zero-shot configurations reach at most 6.00\% at pass@5, and GPT-5.6-Terra-Agent reaches 7.67\%. Additional candidates largely revisit unsuccessful regions of the task set. Few-shot prompting and QASM-Eval adaptation produce much larger improvements, with Llama-70B-ours reaching 33.93\% at pass@1 and 51.33\% at pass@5. Its gain of 17.40 points across the sampling budget is accompanied by continued improvement at later steps, including 2.53 points from the fifth candidate.

Syntax validity becomes high after adaptation while end-to-end success remains much lower. Llama-70B-ours has a syntax-valid rate of 96.27\%, and Llama-8B-ours reaches 90.93\%; their pass@1 rates are 33.93\% and 12.67\%. Complex tasks require several layers to agree simultaneously: a measurement must leave the right conditional state, a controller update must reach the correct actuator, a waveform must inherit the intended frame state, and the resulting events must satisfy the schedule. This combination creates failure opportunities beyond the local construction of a pulse or branch.

The category contrast is pronounced for Llama-8B-ours, whose pulse pass@1 is 83.80\% and complex pass@1 is 12.67\%. Llama-70B-ours improves the complex result to 33.93\%, while its pulse rate is 89.33\%. Model scale within the adapted pair is thus associated with a much larger advantage on the combined workflows. The case studies identify specific dependencies worth strengthening: tracking physical pulse axes through relative phase updates, retaining both free-evolution intervals of an echo, and applying adaptive estimates to accumulated drive phase. These requirements connect the category-level difficulty to observable changes in the generated quantum process.

For a verifier-guided generation system, the five-candidate results indicate useful recovery headroom on complex tasks, alongside a substantial remaining uncovered set. Training and correction examples that expose intermediate state and control dependencies can address these failures directly: the measurement branch state, the frame phase before each pulse, and the event position before a final analysis operation are all concrete quantities a model can be asked to maintain.

\FloatBarrier

\section{Extra Evaluation Results}
\label{appendix:extraresults}

We also evaluated several Qwen-family models on QASM-Eval. However, because \texttt{Qwen3-Coder-480B} was used during dataset construction to generate task variants and TODO prompts, we do not report Qwen-related results in the main text in order to avoid potential bias. 

The Qwen results here are based on the core 100-task test set (25 tasks per category, five generations per task). They are retained for completeness and are not directly comparable with the 1,200-task results in the main text.

For completeness, we provide these results here. Table~\ref{tab:qwen-results} reports the pass@1 scores of \texttt{Qwen3-235B-A22B-Instruct-2507} (\texttt{qwen235b}), \texttt{Qwen3-30B-A3B-Instruct-2507} (\texttt{qwen30b}), and \texttt{Qwen2.5-Coder-7B} (\texttt{qwen7b}).

\begin{center}\begin{minipage}{\linewidth}\centering\captionsetup{type=table}
\centering
\caption{Evaluation results of Qwen-family models on QASM-Eval, measured by pass@1 across task categories.}
\label{tab:qwen-results}
\small
\begin{tabular}{l *{5}{>{\centering\arraybackslash}p{0.13\linewidth}}}
\hline
Model & \multicolumn{5}{c}{pass@1} \\
\cline{2-6}
      & Classical & Timing & Pulse & Complex & Overall \\
\hline
qwen235b & 0.36 & 0.48 & 0.76 & 0.00 & 0.40 \\
qwen30b  & 0.20 & 0.28 & 0.32 & 0.00 & 0.20 \\
qwen7b   & 0.00 & 0.00 & 0.00 & 0.00 & 0.00 \\
\hline
\end{tabular}
\end{minipage}\end{center}

In addition, the error counts for the Qwen family are shown in Figure~\ref{fig:qwen-syntax-breakdown}.

\begin{center}\captionsetup{type=figure}
  \centering
  \includegraphics[width=0.48\linewidth]{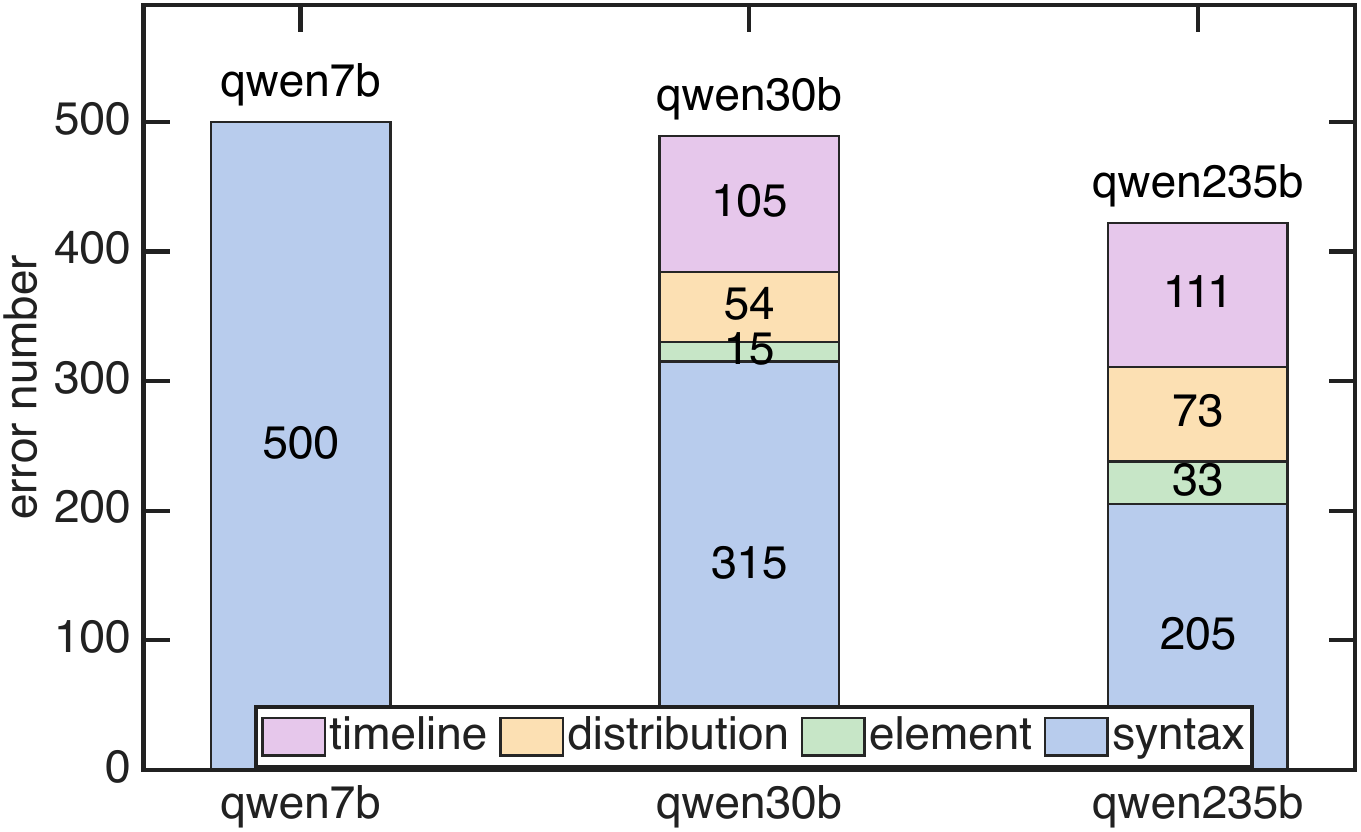}
  \caption{Breakdown of error counts by type for the Qwen family. The evaluation includes 500 samples in total, and each sample may contain multiple errors.}
  \label{fig:qwen-syntax-breakdown}
\end{center}

These results suggest that larger models achieve better overall performance, but their gains remain limited by insufficient mastery of OpenQASM~3 syntax and language-specific constructs.
\FloatBarrier

\FloatBarrier

\section{LLMs for quantum programming}
Large language models (LLMs) have demonstrated strong capabilities on code-generation tasks and have been increasingly adopted in quantum-computing workflows. Early efforts largely focus on constructing gate-level circuits using high-level Python frameworks. A representative example is IBM's Qiskit Code Assistant \citep{dupuis2024qiskit}, which trains LLMs to generate Qiskit programs. Beyond model training, agentic system has also been applied to enhance LLM's Qiskit-generation capability \citep{campbell2025enhancing}. In addition to Qiskit-centric assistants, The agentic framework Pennylang~\citep{basit2025pennylang} explored leveraging retrieval-augmented generation (RAG) in writing Pennylane programs. 

More recently, attention has begun to shift toward hardware-level representations. For example, emerging work explores reinforcement-learning method while training LLMs on OpenQASM programs \citep{yu2025quasar}, as well as attempts to fine-tune LLMs to generate OpenQASM programs for specific set of quantum algorithms \citep{yang2024qcircuitbench}. Despite this progress, existing LLM-for-quantum-programming research either anchors at high-level SDK layer where compilation can hide hardware-facing constraints, or barely cover the advanced features introduced in OpenQASM~3, such as classical logic, timing constraints, and pulse-level control.
\FloatBarrier

\section{Limitations and Future Work} \label{appendix:future}
QASM-Eval provides a reproducible task interface for studying how models use and combine OpenQASM~3 constructs. The following extensions address broader task distributions, test coverage, and device execution.

\paragraph{Generalization to independently written tasks.}
The theme-held-out split tests transfer to task families excluded from training, while expert review and executable references keep the intended behavior clear. Training and test tasks nevertheless share a construction process and description style. An important next step is to collect specifications written independently of reference code, including descriptions from quantum programmers. Evaluating several independently written descriptions of the same physical objective would help distinguish sensitivity to wording from difficulty with the quantum task. Longer programs and open-ended debugging would further test whether the learned skills carry beyond local completion.

\paragraph{Broader verification and expert review.}
The verifier supplies executable acceptance tests for the behavior and constructs specified by each task. As with other execution-based benchmarks, the scope of a pass is determined by these tests. Appendix~\ref{appendix:verifier-fid} provides expert agreement and feature-specific validation. Reference filtering and candidate scoring share an execution stack, and the expert audit covers a finite sample from one model. Broader audits across models and task categories, together with independent implementations, would strengthen this evidence. Additional input states and measurement bases would improve coverage of coherent errors. For tasks that permit several valid schedules, checking the stated timing constraints would allow that flexibility while preserving the intended physical behavior.

\paragraph{Execution on hardware.}
A common virtual device makes the model comparisons reproducible while exercising the classical feedback, timing, and pulse constructs motivated by hardware control. Running these programs on devices also requires backend-specific calibration, resource mapping, and controller-latency constraints. Future work should compile compatible task subsets with provider toolchains and compare their simulated and measured behavior. Testing across devices and calibration settings would clarify which learned control skills transfer and where the generated programs need further adaptation.

\end{document}